\documentclass[11pt]{article}
\usepackage[margin=1in]{geometry}

\usepackage{graphicx}
\usepackage{xcolor}
\usepackage{amssymb}
\usepackage[round]{natbib}

\usepackage{booktabs}
\providecommand{\best}[1]{\textbf{#1}}
\providecommand{\snd}[1]{\underline{#1}}
\providecommand{\trd}[1]{\textit{#1}}

\providecommand{\upsc}[1]{\textup{\scshape #1}}
\providecommand{\grp}[2]{\multicolumn{#1}{@{}l}{\textit{\let\textsc\upsc #2}}}

\usepackage{amsmath}
\usepackage{amsfonts}

\usepackage{algorithm}
\usepackage{algorithmic}

\usepackage{todonotes}
\usepackage{placeins}
\usepackage{verbatim}
\usepackage[hidelinks]{hyperref}
\hypersetup{pdftitle={Potential Games for Self-Neighborhood Channel Mixing in Graph Neural Networks},
 pdfauthor={Marco Lavorini, Davide Bacciu, Giancarlo Bigi}}

\newcommand{\method}{CG2}
\newcommand{\Hmat}{\mathbf{H}}
\newcommand{\R}{\mathbb{R}}

\newtheorem{theorem}{Theorem}

\newtheorem{proposition}{Proposition}

\begin{document}

\title{Potential Games for Self-Neighborhood Channel Mixing in Graph Neural Networks}
\author{Marco Lavorini \and Davide Bacciu \and Giancarlo Bigi}
\date{\normalsize University of Pisa}
\maketitle

\begin{abstract}
   Graph message passing must balance two competing objectives: incorporating neighborhood signals while preserving node-specific information. We model this trade-off as a game in which nodes choose how to mix their own representations with strict-neighborhood aggregates. For fixed channel representations, task supervision defines a common-payoff game, while local distinctness payoffs induce an exact potential game whose potential is the degree-normalized strict Dirichlet energy. We introduce CG2, which amortizes these choices through a shared six-parameter policy per layer based on local channel statistics and can be trained using either objective. Controlled comparisons show that the shared task-trained policy improves mean performance over fixed mixing on three of five heterophily benchmarks, with significant gains on two, while free per-node actions trained with either objective rarely improve on the same baseline. On the ECHO long-range benchmark, the clearest contribution of the gate appears on eccentricity, where the task-trained policy substantially reduces prediction error.
\end{abstract}

\section{Introduction}

Graph Neural Networks (GNNs) based on message passing repeatedly combine a node's current representation with information aggregated from its neighborhood \citep{kipf2017semisupervised,sage,gat}. This mechanism allows information to propagate across the graph, but it also creates a structural tension. Increasing message-passing depth expands the receptive field, yet repeated neighborhood mixing can make node representations increasingly indistinguishable or compress distant signals into fixed-size vectors \citep{Oono2020Graph,alon2021on}. Effective propagation therefore also depends on how each node balances the information of its neighborhood against its own representation, at every layer.

We study this balance through a game-theoretic formulation of self-neighborhood channel mixing. At layer $\ell$, each non-isolated node $i$ is a player whose action $\alpha_i^\ell\in[0,1]$ controls the mixture between a transformed self channel and a transformed strict-neighborhood aggregate. Neighborhood propagation provides the information-sharing component of the interaction, while each node receives a distinctness payoff that rewards local variation in the resulting representations. The action therefore regulates how strongly the node participates in local diffusion.

This formulation yields two structural properties. For fixed channel representations, every unilateral change in a node's payoff equals the corresponding change in the degree-normalized strict Dirichlet energy of the mixed representation, so the layer-local game is an exact potential game. Each node's payoff is also convex in its own action, so wherever the self and neighborhood representations differ, every best response lies in $\{0,1\}$, and pure equilibria assign hard, graph-specific channels. The same players and actions also admit a task-based identical-interest game, with the shared downstream task utility as its potential. The two games therefore place different objectives on the same channel-mixing decisions: a global supervised task potential and a layer-local, label-free distinctness potential.

Rather than computing an equilibrium on every input graph, we amortize the node actions through a shared policy over local channel statistics standardized within each graph. The shared parameters couple the node actions and restrict the admissible action profiles, and the policy is an inductive control rule for nodes and graphs not observed during training. We call the resulting layer Channel-mixing Games on Graphs (\method{}), a minimal two-channel layer whose gate contains only six learnable parameters per layer.

The same policy can play either game: \textsc{TaskGate} is optimized end to end through the task loss, whereas \textsc{GameGate} ascends the layer-local distinctness potential while the remaining parameters follow the task loss. Since the two variants share the forward architecture and policy class, comparing them isolates the effect of the gate-training objective, under their respective training schedules.

\paragraph{Contributions.}
We make four contributions.

\textbf{(i)} We formulate task-based and distinctness-based channel mixing as two potential games over the same players and actions: the task objective, shared by all nodes, and a label-free distinctness potential, the degree-normalized strict Dirichlet energy, whose pure equilibria are hard channel assignments. We use the formulation as a lens on node-wise gating and, through the unilateral regret of trained gates, as a diagnostic.

\textbf{(ii)} We introduce \method{}, which plays either game through a shared six-parameter policy per layer over local channel statistics, an inductive control rule for nodes and graphs not seen during training.

\textbf{(iii)} We compare the shared policy with fixed mixing, free per-node actions, and a frozen random policy. The task-trained policy improves on fixed mixing by one to two points on three of five datasets, significantly on all three with five seeds per split, whereas free per-node actions rarely improve on it. Varying the actions across nodes accounts for part of this gain and learning adds to it, both significantly on Roman-empire. On the heterophily suite, a degree-dependent self weight recovers this gain, as a two-parameter gate that sees only the degree performs as well. Of the two training signals, the task loss is the more effective one for the policy.

\textbf{(iv)} Following \citet{luo2024classic}, we also give four channel mixers residual connections and batch normalization, which improve them on almost every dataset, as they improve classic GNNs; with these components, \method{} remains competitive across the heterophily suite and is significantly ahead of tuned classic backbones on Roman-empire. On the ECHO long-range benchmark, the gate lowers the eccentricity error of fixed mixing by about a third.
\section{Related Work}
\label{sec:related}
\paragraph{Heterophily and long-range propagation.}
Heterophily and long-range propagation challenge message passing in different ways, but both make it important to regulate how neighborhood information is incorporated into node representations. In heterophilic graphs, adjacent nodes may have dissimilar labels or features, making indiscriminate local aggregation unreliable. This has motivated methods that adapt local propagation or incorporate broader structural information \citep{yan2022two,li2022findingglobalhomophilygraph}. Evaluation in this
regime is also sensitive to dataset quality, split protocols, and baseline tuning \citep{platonov2024criticallookevaluationgnns,luo2024classic}.

Long-range tasks pose a different difficulty: information from distant nodes must remain influential after multiple propagation steps. Increasing depth expands the receptive field, but deeper propagation can homogenize node representations, while distant information may be attenuated by computational or topology-induced bottlenecks \citep{rusch2023surveyoversmoothinggraphneural, digiovanni2023oversquashingmessagepassingneural, arnaizrodriguez2026oversmoothing}.
A-DGN, DRew, and SONAR modify the propagation dynamics through stable updates, delayed communication, and wave-based propagation, respectively \citep{gravina2023antisymmetric,drew,trenta2025sonar}. ECHO evaluates long-range propagation through controlled tasks \citep{miglior2026can}. We investigate whether node-wise control of self-neighborhood mixing provides a minimal mechanism that remains useful across both regimes.

\paragraph{Self-neighborhood channel mixing.}
A related line of work controls how different sources of graph information are combined. H2GCN keeps self and neighbor representations separate and combines information from multiple neighborhood ranges \citep{h2gcn}. GPR-GNN, FAGCN, ACM-GCN, GATE, G$^2$, and CHAT-GNN make propagation adaptive through coefficients over propagation steps, signed edge weights, node-wise filter mixing, separate self-loop attention, gradient-gated update rates, or channel attention \citep{chien2021adaptive,bo2021beyond,luan2022revisiting,mustafa2024gateintrusiveneighbors,rusch2023gradientgating,chatgnn2024}.

Among these methods, ACM-GCN, GATE, and G$^2$ are the closest comparisons in terms of node-wise adaptivity. ACM-GCN learns node-specific weights over identity, aggregation, and diversification channels. GATE scores the self-loop with its own attention parameters, so that each node can weight itself against its neighbors and switch off neighborhood aggregation that does not benefit the task. G$^2$ gates each node's update, per channel, by local graph gradients, so that nodes stop updating as their neighborhoods become uniform. \method{} instead uses only two transformed channels, the self-representation and the strict-neighborhood aggregate, and produces one node-wise mixing coefficient through a shared six-parameter policy over local channel statistics. GCNH is closer in its forward structure: it also separately transforms the center node and its neighborhood, but combines them using one learned coefficient per layer rather than node-wise coefficients \citep{10191196}. In \method{}, the node-wise coefficients are the players' actions in a layer-local potential game.

\paragraph{Game-theoretic graph learning.}
Game theory has been used in graph learning both for explanation and for modeling propagation. GStarX and a game-theoretic counterfactual method explain predictions through cooperative games and semivalues \citep{zhang2022explaining,chhablani2024gametheoretic}. Closer to propagation, Game-of-GNN models graph representation learning as a mean-field game and learns graph-dependent message-passing operators~\citep{dan2025explore}, and Co-GNN treats each node as a player that chooses whether to listen, broadcast, both, or neither, so that an action network selects the message-passing graph at every layer \citep{pmlr-v235-finkelshtein24a}. In \method{}, graph nodes instead choose scalar actions that control self-neighborhood mixing within a single layer, and the game, with explicit utilities and an exact potential, supplies a training signal for these actions rather than an explanation or a model of the full message-passing dynamics. For fixed channel representations, the potential is the degree-normalized strict Dirichlet energy, a measure of variation in graph representations \citep{zhou2021dirichlet,digiovanni2023understanding}.

\section{Self-Neighborhood Channel Mixing}\label{sec:method}

Let $G=(V,E)$ be an undirected graph with node set $V$, strict neighborhood $\mathcal{N}'(i)=\{j:(i,j)\in E,\ j\neq i\}$, and strict degree $d_i=|\mathcal{N}'(i)|$. We write $V_{+}=\{i\in V : d_i>0\}$ for the set of non-isolated nodes. Let $\Hmat^\ell \in \R^{|V|\times p_\ell}$ denote the node states at layer $\ell$, where $p_\ell$ is the feature dimension. A \method{} layer computes two channels:
\begin{equation}
\begin{aligned}
  \mathbf{x}_{i,\mathrm{self}}^\ell
  &=
  \mathbf{W}_{\mathrm{self}}^\ell
  \mathbf{h}_i^\ell, \\
  \mathbf{m}_i^\ell
  &=
  \Phi_{\mathbf{W}_{\mathrm{neigh}}^\ell}
  \!\left(
    \mathbf{H}^\ell,
    \mathcal{N}'(i)
  \right)
\end{aligned}
\label{eq:channels}
\end{equation}
Here,
$\mathbf{W}_{\mathrm{self}}^\ell$ and $\mathbf{W}_{\mathrm{neigh}}^\ell$ are independent learnable transformations. The neighborhood operator
$\Phi_{\mathbf{W}_{\mathrm{neigh}}^\ell}$ is defined by
\begin{equation}
  \Phi_{\mathbf{W}_{\mathrm{neigh}}^\ell}
  \!\left(\mathbf{H}^\ell,\mathcal{N}'(i)\right)
  :=
  \operatorname{AGG}_{j\in\mathcal{N}'(i)}^\ell
  \!\left(
    \mathbf{W}_{\mathrm{neigh}}^\ell
    \mathbf{h}_j^\ell
  \right).
  \label{eq:aggregation}
\end{equation}
The operator $\operatorname{AGG}^\ell$ is permutation invariant with respect to the ordering of the neighbors. In our experiments, it is instantiated using either sum or mean aggregation, depending on the task and configuration. The term \emph{strict} indicates that graph node $i$ is excluded from its neighborhood channel; it does not impose a particular aggregation or degree normalization.

The two channels are combined through a node- and
layer-specific coefficient $\alpha_i^\ell\in[0,1]$:
\begin{equation}
  \mathbf{r}_i^\ell
  =
  \alpha_i^\ell
  \mathbf{x}_{i,\mathrm{self}}^\ell
  +
  \left(1-\alpha_i^\ell\right)
  \mathbf{m}_i^\ell .
  \label{eq:mix}
\end{equation}
The coefficient $\alpha_i^\ell$ is the self-channel weight. Large values favor the transformed self-representation, while small values favor strict-neighborhood propagation.

For an isolated node $i\notin V_{+}$, we set $\mathbf{m}_i^\ell=\mathbf{0}$ and $\alpha_i^\ell=1$, so $\mathbf{r}_i^\ell=\mathbf{x}_{i,\mathrm{self}}^\ell$ and the gate is bypassed.

\begin{figure*}[t]
    \centering
    \includegraphics[width=0.75\textwidth]{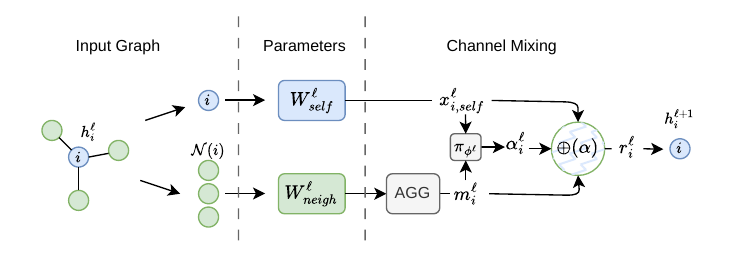}
    \caption{Overview of the \method{} layer. Separate transformed self and strict-neighborhood channels are combined by the node- and layer-specific gate $\alpha_i^\ell$, and the mixed representation is passed to the backbone block.}
    \label{fig:layer}
\end{figure*}

The mixed representation is then passed to the backbone block, $\mathbf{h}_i^{\ell+1}=\mathcal{B}^\ell(\mathbf{r}_i^\ell,\mathbf{h}_i^\ell)$, where $\mathcal{B}^\ell$ denotes the chosen combination of residual connections, normalization, activation, and dropout inherited from the tuned backbone \citep{he2015deepresiduallearningimage, ioffe2015batchnormalizationacceleratingdeep, ba2016layer, luo2024classic}. Thus, \method{} changes how the self and neighborhood channels are combined while leaving the remaining backbone structure unchanged. Figure~\ref{fig:layer} summarizes the layer.

\section{A Layer-Local Potential Game}
\label{sec:game}

We now formalize the game-theoretic interpretation of Eq.~\eqref{eq:mix}.  Fix a layer $\ell$, its input states $\mathbf{H}^\ell$, and its
channel transformations. With these quantities fixed, $\mathbf{x}_{i,\mathrm{self}}^\ell$ and $\mathbf{m}_i^\ell$ are fixed with respect to the gate actions. The players are the non-isolated graph nodes $i\in V_+$. Each player chooses a continuous action
\[
\alpha_i^\ell\in\mathcal{A}_i=[0,1],
\]
where $\alpha_i^\ell=1$ selects the transformed self channel, $\alpha_i^\ell=0$ selects the strict-neighborhood channel, and intermediate actions deterministically interpolate between them. The action profile
\[
\boldsymbol{\alpha}^\ell
=
\left(\alpha_i^\ell\right)_{i\in V_+}
\]
determines the mixed representations
\[
\mathbf{r}_i^\ell
=
\alpha_i^\ell\mathbf{x}_{i,\mathrm{self}}^\ell
+
\left(1-\alpha_i^\ell\right)\mathbf{m}_i^\ell,
\]
collected in $\mathbf{R}^\ell$.

The neighborhood channel provides the information-sharing component of message passing. We complement it with a distinctness payoff that opposes homogenization: each node is rewarded for keeping its mixed representation distinct from those of its strict neighbors.

Define the degree-normalized strict Dirichlet energy
\begin{equation}
\mathcal{E}^\ell
\left(
  \boldsymbol{\alpha}^\ell
\right)
=
\frac{1}{2|V|}
\sum_{i\in V_+}
\frac{1}{d_i}
\sum_{j\in\mathcal{N}'(i)}
\left\|
  \mathbf{r}_i^\ell
  -
  \mathbf{r}_j^\ell
\right\|_2^2 ,
\label{eq:dirichlet}
\end{equation}
and assign graph node $i$ the utility
\begin{equation}
u_i^\ell
\left(
  \boldsymbol{\alpha}^\ell
\right)
=
\frac{1}{2|V|}
\sum_{j\in\mathcal{N}'(i)}
\left(
  \frac{1}{d_i}
  +
  \frac{1}{d_j}
\right)
\left\|
  \mathbf{r}_i^\ell
  -
  \mathbf{r}_j^\ell
\right\|_2^2 .
\label{eq:utility}
\end{equation}

\begin{theorem}[Exact potential]
\label{thm:potential}
For fixed channel values, the game
\[
\mathcal{G}^\ell
=
\left(
  V_+,
  \{\mathcal{A}_i\}_{i\in V_+},
  \{u_i^\ell\}_{i\in V_+}
\right)
\]
is an exact potential game \citep{monderer1996potential} with potential $\mathcal{E}^\ell$. In particular, for every node $i\in V_+$ and every unilateral deviation $\alpha_i^\ell\rightarrow{\alpha_i^\ell}'$,
\begin{align}
&
u_i^\ell
\left(
  {\alpha_i^\ell}',
  \boldsymbol{\alpha}_{-i}^\ell
\right)
-
u_i^\ell
\left(
  \alpha_i^\ell,
  \boldsymbol{\alpha}_{-i}^\ell
\right)
\nonumber\\
&\qquad =
\mathcal{E}^\ell
\left(
  {\alpha_i^\ell}',
  \boldsymbol{\alpha}_{-i}^\ell
\right)
-
\mathcal{E}^\ell
\left(
  \alpha_i^\ell,
  \boldsymbol{\alpha}_{-i}^\ell
\right).
\label{eq:potential_identity}
\end{align}
\end{theorem}

\noindent\emph{Proof.}
Because the layer inputs and channel values are fixed, changing $\alpha_i^\ell$ modifies only $\mathbf{r}_i^\ell$. In Eq.~\eqref{eq:dirichlet}, every edge incident to graph node $i$ appears twice: once in the term indexed by $i$, with weight $1/d_i$, and once in the term indexed by its neighbor $j$, with weight $1/d_j$. The part of the energy affected by the unilateral deviation is therefore exactly Eq.~\eqref{eq:utility}. All non-incident terms cancel in the difference.
\hfill$\square$

Theorem~\ref{thm:potential} holds for any backbone block and any neighborhood aggregation. The energy is evaluated on $\mathbf{R}^\ell$ before applying the backbone block $\mathcal{B}^\ell$, so that a unilateral action change modifies only $\mathbf{r}_i^\ell$ even when the backbone couples graph nodes, as batch normalization does, and the argument requires only that the channel values be fixed with respect to the gate actions.

By Eq.~\eqref{eq:utility}, the utility of node $i$ depends, through the shared edge terms, on the actions of its strict neighbors: the graph specifies who interacts with whom, and each node's best response depends on its neighbors' actions. Formally, given the actions $\boldsymbol{\alpha}_{-i}^\ell$ of the other nodes, a best response of node $i$ is any maximizer of $u_i^\ell(\cdot,\boldsymbol{\alpha}_{-i}^\ell)$ over $[0,1]$, and a pure Nash equilibrium is an action profile in which every node plays a best response. We call node $i$ nondegenerate if $\mathbf{x}_{i,\mathrm{self}}^\ell\neq\mathbf{m}_i^\ell$.

\begin{proposition}[Endpoint best responses]
\label{prop:endpoint}
For fixed channel values and any actions of the other nodes, every best response of a nondegenerate node lies in $\{0,1\}$, whereas every action in $[0,1]$ is a best response at a degenerate node. The game $\mathcal{G}^\ell$ admits a pure Nash equilibrium in $\{0,1\}^{|V_+|}$, and every pure Nash equilibrium assigns endpoint actions to all nondegenerate nodes.
\end{proposition}

\noindent\emph{Proof.}
Fix $i\in V_+$ and $\boldsymbol{\alpha}_{-i}^\ell$. Since
\[
\mathbf{r}_i^\ell
=
\mathbf{m}_i^\ell
+
\alpha_i^\ell
\left(
  \mathbf{x}_{i,\mathrm{self}}^\ell
  -
  \mathbf{m}_i^\ell
\right),
\]
the utility in Eq.~\eqref{eq:utility} is a positively weighted sum of squared affine functions of $\alpha_i^\ell$. It is therefore strictly convex at nondegenerate nodes and constant at degenerate nodes, which gives the stated best responses.

The continuous potential $\mathcal{E}^\ell$ attains a maximum on the compact action space $[0,1]^{|V_+|}$. By Theorem~\ref{thm:potential}, every maximizer is a pure Nash equilibrium and therefore uses endpoint actions at all nondegenerate nodes. Replacing the remaining actions by endpoints changes no mixed representation or utility, yielding a pure Nash equilibrium in $\{0,1\}^{|V_+|}$.
\hfill$\square$

\paragraph{The task game.}
The task objective defines a second game on the same players and actions. Consider varying only the actions at layer $\ell$ while leaving the rest of the network unchanged, so that the downstream task loss can be viewed as a function of $\boldsymbol{\alpha}^\ell$. Assigning each node $i\in V_+$ the common utility $u_{i,\mathrm{task}}^\ell(\boldsymbol{\alpha}^\ell):=-\mathcal{L}_{\mathrm{task}}(\boldsymbol{\alpha}^\ell)$ defines a common-payoff, or identical-interest, game \citep{identical}. The common payoff is itself an exact potential \citep{monderer1996potential}, since any unilateral deviation produces the same change in the player's utility and in the potential. The two games therefore share players and action sets but differ in their potentials. The task potential is global: changing one action can propagate through the remaining network and affect the overall task loss. It is also generally nonconvex in the actions, so Proposition~\ref{prop:endpoint} does not apply. The distinctness potential is local: under the fixed-channel setting of Theorem~\ref{thm:potential}, changing $\alpha_i^\ell$ changes only $\mathbf{r}_i^\ell$, and hence affects $\mathcal{E}^\ell$ only through terms involving node $i$ and its strict neighbors.

Proposition~\ref{prop:endpoint} characterizes the equilibria of the distinctness game, in which, as in the task game, every node independently controls its own action. We seek instead an inductive control rule that sets each node's balance between self and neighborhood information from local channel statistics, and amortize the action profile through a shared, low-capacity policy. 

\paragraph{Amortized action policy.}
The shared policy maps a five-dimensional descriptor of the two channels to the node action. For each non-isolated node, we define
\begin{align}
\mathbf{q}_i^\ell =
\Big[
&
\log(1+d_i),\;
\|\mathbf{x}_{i,\mathrm{self}}^\ell\|_2,\;
\|\mathbf{m}_i^\ell\|_2,
\nonumber\\
&
\|\mathbf{x}_{i,\mathrm{self}}^\ell-\mathbf{m}_i^\ell\|_2,\;
\operatorname{cosim}
\left(
  \mathbf{x}_{i,\mathrm{self}}^\ell,
  \mathbf{m}_i^\ell
\right)
\Big]^\top .
\label{eq:descriptor}
\end{align}
Here, $\operatorname{cosim}$ denotes cosine similarity, computed with a small denominator clamp for numerical stability. The descriptor records the node degree, the magnitudes of the two channels, and their Euclidean and angular disagreement. Each coordinate is standardized with its mean and standard deviation over the nodes of the same input graph, so the policy reads each node relative to the rest of its graph. The standardized descriptor is detached before being passed to the policy.

The action is produced by the policy
\begin{equation}
\alpha_i^\ell
=
\pi_{\boldsymbol{\phi}^\ell}
\left(
  \mathbf{q}_i^\ell
\right)
=
\sigma
\left(
  (\mathbf{w}^\ell)^\top
  \mathbf{q}_i^\ell
  +
  b^\ell
\right),
\label{eq:policy}
\end{equation}
where $\boldsymbol{\phi}^\ell=(\mathbf{w}^\ell,b^\ell)$ denotes the learnable parameters shared across all nodes at layer $\ell$. Since the descriptor has five components, the policy introduces only six parameters per layer. Most of the layer's trainable capacity therefore remains in the self and neighborhood transformations, while the policy controls their node-wise mixture.

\paragraph{Two training signals for the same policy.}
We use the policy in Eq.~\eqref{eq:policy} in two variants.
\textup{TaskGate} trains the policy through the downstream task loss,
equivalently maximizing the potential of the task game.
\textup{GameGate} uses the same forward map and parameterization,
but trains the policy by ascending the layer potential in
Eq.~\eqref{eq:dirichlet}. The two variants therefore isolate the
effect of the gate-training signal while keeping the policy architecture
and action space fixed.

In both \textup{TaskGate} and \textup{GameGate}, the node actions are
not optimized as independent variables; instead, they are generated
by the shared policy in Eq.~\eqref{eq:policy}. For fixed channel values
and descriptors, the gradient of either potential with respect to the
shared policy parameters aggregates the players' marginal payoffs
through the policy Jacobian. Parameter sharing restricts the available
joint action changes, so stationarity of the shared policy does not
imply individual optimality or a Nash equilibrium.
Proposition~\ref{prop:amortized} in Appendix~\ref{app:theory}
formalizes this distinction.

\subsection{Optimization and Training Schedules}
\label{sec:training}

Let $\boldsymbol{\theta}$ denote all non-gate parameters and $\boldsymbol{\phi}$ the gate-policy parameters.

\paragraph{TaskGate.}
The task-trained variant is optimized end to end through
\begin{equation}
  \min_{\boldsymbol{\theta},\boldsymbol{\phi}}
  \mathcal{L}_{\mathrm{task}}
  \left(
    \boldsymbol{\theta},
    \boldsymbol{\phi}
  \right).
  \label{eq:task-objective}
\end{equation}
A single backward pass updates both the non-gate parameters and the gate policies. This optimizes the potential of the task game, $-\mathcal{L}_{\mathrm{task}}$, over the action profiles generated by the shared policy.

\paragraph{GameGate.}
GameGate trains the gate policies separately from the remaining network parameters. For each layer $\ell$, the gate policy is optimized through
\begin{equation}
  \mathcal{L}_{\mathrm{game}}^\ell(\boldsymbol{\phi}^\ell)
  =
  -\mathcal{E}^\ell(\mathbf{R}^\ell).
  \label{eq:game-loss}
\end{equation}
When computing this loss, the layer input $\mathbf{H}^\ell$ is replaced by $\operatorname{sg}(\mathbf{H}^\ell)$, where stop-gradient leaves its argument unchanged in the forward pass but blocks gradients in the backward pass. With the non-gate parameters fixed, minimizing Eq.~\eqref{eq:game-loss} therefore maximizes the layer potential for the current channel representations.

Training begins with $T_{\mathrm{init}}$ task-only optimization steps with the gate policies frozen. It then alternates $n_{\mathrm{game}}$ updates of the layer-wise gate policies through their game losses with $n_{\mathrm{task}}$ updates of $\boldsymbol{\theta}$ through $\mathcal{L}_{\mathrm{task}}$, as summarized in Algorithm~\ref{alg:gamegate}.

\begin{algorithm}[tb]
\caption{Training schedule for \textsc{GameGate}.}
\label{alg:gamegate}
\begin{algorithmic}[1]

\REQUIRE Training data $\mathcal{D}$, warm-up length
$T_{\mathrm{init}}$, alternating rounds $T$, updates per round $n_{\mathrm{game}}$ and $n_{\mathrm{task}}$

\FOR{$t=1,\ldots,T_{\mathrm{init}}$}
    \STATE With $\boldsymbol{\phi}$ fixed, update
    $\boldsymbol{\theta}$ through $\mathcal{L}_{\mathrm{task}}$
\ENDFOR

\FOR{$t=1,\ldots,T$}
    \STATE With $\boldsymbol{\theta}$ fixed, update each
    $\boldsymbol{\phi}^{\ell}$ through
    $\mathcal{L}_{\mathrm{game}}^{\ell}$, $n_{\mathrm{game}}$ times
    \STATE With $\boldsymbol{\phi}$ fixed, update
    $\boldsymbol{\theta}$ through $\mathcal{L}_{\mathrm{task}}$, $n_{\mathrm{task}}$ times
\ENDFOR

\STATE \textbf{return} $\boldsymbol{\theta},\boldsymbol{\phi}$

\end{algorithmic}
\end{algorithm}
\section{Experiments}
\label{sec:experiments}

We evaluate on the five heterophily datasets of \citet{platonov2024criticallookevaluationgnns}, Roman-empire, Amazon-ratings, Minesweeper, Questions, and Tolokers, with their ten official splits, and on the long-range ECHO benchmark \citep{miglior2026can}.

\subsection{Setup}
\label{sec:setup}

On the heterophily datasets, every model receives 500 task-training updates and is selected by mean validation performance within its own hyperparameter search space. \method{} and the classic GNNs use residual connections and batch normalization following \citet{luo2024classic}, whereas the channel-mixer implementations retain their original architectural form. \textsc{GameGate} follows Algorithm~\ref{alg:gamegate}, with additional gate updates but the same number of task updates as the other models.

On ECHO, our runs share depth, width, learning rate, and task-specific training budgets rather than following the benchmark's model-specific hyperparameter search. \method{} and the classic backbones additionally use residual connections, batch normalization, and a jumping-knowledge readout. Validation selects mean or sum aggregation for each \method{} variant and task.

We report the test score at the checkpoint with the best validation performance, as mean and standard deviation over ten heterophily splits or five ECHO seeds. Paired heterophily comparisons use two-sided Wilcoxon signed-rank tests, with Holm correction within the families specified in Appendix~\ref{app:stats}. Appendices~\ref{app:setup} and~\ref{app:echo} give the full setup.

\subsection{Heterophily and long-range benchmarks}
\label{sec:benchmarks}

\begin{table*}[t]
\centering
\caption{Heterophily benchmarks: mean $\pm$ standard deviation over the ten official splits (accuracy on Roman-empire and Amazon-ratings, ROC-AUC on Minesweeper, Questions, and Tolokers). Best classic is selected by validation from GCN, GraphSAGE, and GAT, run on the configuration grid of \method{} with the same residual connections and batch normalization. Bold, underline, and italics mark the three best values per column.}
\label{tab:hetero}
\small
\begin{tabular}{@{}lccccc@{}}
\toprule
Model & Roman-emp. & Amazon-rat. & Minesw. & Questions & Tolokers \\
\midrule
\grp{6}{Ours (\method{})}\\
\textsc{TaskGate} & \best{89.2$\pm$0.4} & \snd{54.0$\pm$0.3} & \snd{95.4$\pm$0.4} & \snd{77.0$\pm$2.1} & \snd{85.1$\pm$0.5} \\
\textsc{GameGate} & \trd{88.4$\pm$0.8} & \trd{53.9$\pm$0.5} & \snd{95.4$\pm$0.7} & 75.6$\pm$1.5 & \trd{84.6$\pm$0.7} \\
\midrule
\grp{6}{Channel mixers}\\
FAGCN & 69.4$\pm$0.4 & 44.0$\pm$0.4 & 88.6$\pm$0.9 & 73.6$\pm$1.5 & 80.0$\pm$0.9 \\
GPR-GNN & 73.2$\pm$0.6 & 47.8$\pm$0.6 & 89.5$\pm$0.6 & 74.2$\pm$1.1 & 74.4$\pm$1.0 \\
ACM-GCN & 85.3$\pm$0.7 & 45.7$\pm$0.6 & 90.1$\pm$0.6 & 76.2$\pm$0.9 & 79.3$\pm$1.0 \\
H2GCN & 80.3$\pm$0.5 & 47.7$\pm$0.4 & 90.3$\pm$0.5 & 76.6$\pm$1.2 & 81.2$\pm$0.9 \\
CHAT-GNN & \snd{88.5$\pm$0.6} & 51.5$\pm$0.2 & \trd{92.7$\pm$0.7} & \trd{76.8$\pm$1.5} & 83.6$\pm$1.0 \\
GCNH & 81.3$\pm$0.5 & 46.7$\pm$0.4 & 91.5$\pm$0.6 & 74.8$\pm$0.9 & 84.3$\pm$0.8 \\
GATE & 75.6$\pm$0.4 & 48.6$\pm$0.4 & 87.9$\pm$0.6 & 74.3$\pm$0.9 & 82.3$\pm$0.7 \\
\midrule
Best classic & \snd{88.5$\pm$0.4} & \best{54.1$\pm$0.6} & \best{96.4$\pm$0.4} & \best{77.1$\pm$0.8} & \best{85.6$\pm$0.8} \\
\bottomrule
\end{tabular}

\end{table*}

\begin{table}[t]
\centering
\caption{ECHO-Synth test MAE in the benchmark's target scaling (lower is better) over five seeds. \textsc{Fixed} sets every action to $1/2$, \textsc{Task-MLP} uses an MLP policy, and best classic is the validation-selected GCN, GraphSAGE, or GAT. Bold, underline, and italics mark the three best of our runs. $^{\dagger}$Quoted from \citet{miglior2026can}, under its model-specific tuning.}
\label{tab:echo_synth}
{\footnotesize\setlength{\tabcolsep}{3pt}%
\begin{tabular}{lccc}
\toprule
Model & SSSP & ECC & Diam \\
\midrule
\grp{4}{Ours (\method{})}\\
\textsc{TaskGate} & \best{0.63$\pm$0.03} & \snd{3.30$\pm$0.15} & \trd{1.52$\pm$0.16} \\
\textsc{Task-MLP} & \best{0.63$\pm$0.05} & \best{3.21$\pm$0.12} & \snd{1.45$\pm$0.13} \\
\textsc{GameGate} & \trd{0.72$\pm$0.15} & \trd{4.35$\pm$0.44} & 1.59$\pm$0.08 \\
\textsc{Fixed} & \trd{0.72$\pm$0.07} & 5.15$\pm$0.09 & 2.56$\pm$0.26 \\
\midrule
\grp{4}{Baselines (our runs)}\\
ACM-GCN & \snd{0.68$\pm$0.06} & 4.78$\pm$0.04 & \best{1.26$\pm$0.06} \\
FAGCN & 2.39$\pm$0.23 & 5.22$\pm$0.02 & 2.24$\pm$0.11 \\
GPR-GNN & 2.86$\pm$0.05 & 5.33$\pm$0.01 & 3.52$\pm$0.07 \\
H2GCN & 4.50$\pm$0.02 & 5.33$\pm$0.01 & 2.65$\pm$0.40 \\
CHAT-GNN & 2.15$\pm$0.94 & 5.03$\pm$0.01 & 1.93$\pm$1.18 \\
GATE & 0.87$\pm$0.12 & 5.56$\pm$0.00 & 3.78$\pm$1.37 \\
Best classic & 1.23$\pm$0.28 & 5.14$\pm$0.17 & 1.80$\pm$0.28 \\
\midrule
\grp{4}{References}\\
GCN$^{\dagger}$ & 2.10 & 5.23 & 3.83 \\
A-DGN$^{\dagger}$ & 1.18 & 4.98 & 1.15 \\
Leaderboard best$^{\dagger}$ & 0.12 & 4.65 & 1.01 \\
\bottomrule
\end{tabular}}

\end{table}

\begin{table}[t]
\centering
\caption{ECHO-Chem test MAE (lower is better) over five seeds; charge in units of $10^{-3}$. Among our runs, only the main \method{} rows use bond attributes, so no entry is ranked; Appendix~\ref{app:echo} reports the charge controls without them. Best classic is the validation-selected GCN, GraphSAGE, or GAT. $^{\dagger}$Quoted from \citet{miglior2026can}, under its model-specific tuning.}
\label{tab:echo_chem}
{\footnotesize\setlength{\tabcolsep}{5pt}%
\begin{tabular}{lcc}
\toprule
Model & Charge & Energy \\
\midrule
\grp{3}{Ours (\method{})}\\
\textsc{TaskGate} & 5.66$\pm$0.06 & 19.90$\pm$5.11 \\
\textsc{Task-MLP} & 5.56$\pm$0.15 & 25.52$\pm$6.69 \\
\textsc{GameGate} & 5.88$\pm$0.14 & 21.79$\pm$6.67 \\
\textsc{Fixed} & 5.69$\pm$0.09 & 24.34$\pm$6.73 \\
\midrule
\grp{3}{Baselines (our runs)}\\
ACM-GCN & 7.72$\pm$0.11 & 23.15$\pm$6.56 \\
FAGCN & 14.68$\pm$0.57 & 19.09$\pm$4.80 \\
GPR-GNN & 15.46$\pm$0.44 & 8.06$\pm$1.59 \\
H2GCN & 12.19$\pm$0.18 & 14.73$\pm$5.23 \\
CHAT-GNN & 7.08$\pm$0.06 & 21.64$\pm$4.50 \\
GATE & 46.57$\pm$10.63 & 41.35$\pm$3.74 \\
Best classic & 6.97$\pm$0.07 & 18.06$\pm$5.75 \\
\midrule
Leaderboard best$^{\dagger}$ & 6.11 & 5.26 \\
\bottomrule
\end{tabular}}

\end{table}

\paragraph{Heterophily.}

\textsc{TaskGate} is ahead of every channel mixer in Table~\ref{tab:hetero}, but residual connections and batch normalization account for most of this margin: as for classic GNNs \citep{luo2024classic}, adding them to four mixers brings the strongest within half a point of \textsc{TaskGate}, or above it, on every dataset (Appendix~\ref{app:hetero}). Against classic backbones with the same components and grid, \textsc{TaskGate} leads significantly on Roman-empire, matches them on Amazon-ratings and Questions, and is significantly behind only on Minesweeper. At the longer published configurations of \citet{luo2024classic}, it stays significantly ahead on Roman-empire, where it also improves on fixed mixing, while the classic backbones lead on the other datasets run there (Appendix~\ref{app:hetero}). Since this comparison depends so much on the backbone, we measure the gate against fixed mixing on the same backbone.

\paragraph{Long-range tasks.}

On ECHO (Tables~\ref{tab:echo_synth} and~\ref{tab:echo_chem}), the gate helps most on the synthetic tasks. \textsc{TaskGate} substantially lowers the error of fixed mixing on eccentricity and diameter. Among our runs, \method{} has the lowest error on eccentricity and shortest paths, and ACM-GCN on diameter. The molecular tasks give a weaker picture: \textsc{TaskGate} and \textsc{Fixed} are close on charge, and several baselines, most clearly GPR-GNN and H2GCN, have lower errors on energy. Under their model-specific tuning, the best quoted references remain stronger on shortest paths, diameter, and energy.

\subsection{Free versus amortized strategies}
\label{sec:free}

\begin{table}[t]
\centering
\caption{Free per-node actions and the shared policy, trained on the task loss or the distinctness potential: mean $\pm$ standard deviation over the ten splits. Free actions use the backbone configuration of the corresponding policy, with optimizer settings selected by validation (Appendix~\ref{app:free}). The transferred row freezes the \textsc{TaskGate} policy of each other dataset in the target's configuration, averaged over the four sources (Appendix~\ref{app:transfer}). \textsc{Fixed} is selected by validation over the eight configurations (Appendix~\ref{app:ladder}).}
\label{tab:free}
{\scriptsize\setlength{\tabcolsep}{2pt}%
\begin{tabular}{@{}lccccc@{}}
\toprule
Actions & Rom. & Amz. & Min. & Que. & Tol. \\
\midrule
\textsc{Fixed} & \trd{87.3$\pm$0.6} & 53.8$\pm$0.5 & \snd{95.4$\pm$0.4} & \trd{75.2$\pm$1.3} & 84.1$\pm$0.6 \\
\midrule
\grp{6}{Task game}\\
free actions & 87.0$\pm$0.5 & \best{54.1$\pm$0.4} & \best{97.3$\pm$0.6} & 74.5$\pm$1.0 & 84.0$\pm$0.9 \\
\textsc{TaskGate} & \best{89.2$\pm$0.4} & \snd{54.0$\pm$0.3} & \snd{95.4$\pm$0.4} & \best{77.0$\pm$2.1} & \best{85.1$\pm$0.5} \\
transferred & 88.5$\pm$0.3 & 53.7$\pm$0.4 & 95.3$\pm$0.5 & 76.0$\pm$0.8 & \trd{84.5$\pm$0.5} \\
\midrule
\grp{6}{Distinctness game}\\
free actions & 87.1$\pm$0.5 & \best{54.1$\pm$0.4} & \trd{95.2$\pm$0.5} & 74.1$\pm$1.3 & \trd{84.5$\pm$0.6} \\
\textsc{GameGate} & \snd{88.4$\pm$0.8} & \trd{53.9$\pm$0.5} & \snd{95.4$\pm$0.7} & \snd{75.7$\pm$1.6} & \snd{84.6$\pm$0.7} \\
\bottomrule
\end{tabular}}

\end{table}

\paragraph{Free and shared actions.}

Table~\ref{tab:free} asks whether node actions need a shared policy. Under the task loss, the shared policy improves on fixed mixing by one to two points on Roman-empire, Questions, and Tolokers, also when fixed mixing uses the policy's configuration, whereas free per-node actions stay close to fixed mixing there. Free actions gain only on Minesweeper, a transductive advantage that gives no rule for unseen nodes or graphs. Under the distinctness potential, the shared policy is again ahead of free actions on Roman-empire and Questions and close to them elsewhere. In the mixing ladder (Appendix~\ref{app:ladder}), one coefficient per layer falls behind the node-wise policy wherever the gate helps and is ahead of it on Minesweeper, and the larger \textsc{Task-MLP} policy performs on par, so six parameters per layer suffice.

\paragraph{What the policy learns.}

Part of the gain comes from varying the actions across nodes at all. At the selected \textsc{TaskGate} configuration, a frozen random policy, whose actions still depend on the trained channels, already improves on fixed mixing where the gate helps, and lowers the score on Amazon-ratings and Minesweeper. Learning adds to this. With five seeds per split, the learned policy beats fixed mixing on each dataset where the gate helps; on Roman-empire, the random policy improves significantly on fixed mixing, and the learned policy on the random one (Appendix~\ref{app:random_policy}). A policy learned on another dataset and frozen in the target keeps, on average over the sources, about as much of the gain as the random policy, without its losses on Amazon-ratings and Minesweeper (Appendix~\ref{app:transfer}).

\begin{figure}[tb]
\centering
\includegraphics[width=\columnwidth]{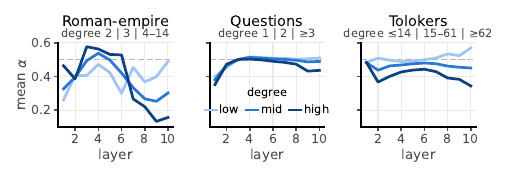}
\caption{Mean action of the \textsc{TaskGate} policy at each layer for the nodes of low, middle, and high strict degree, split at its 1/3 and 2/3 quantiles (ranges under each dataset name), averaged over the checkpoints of splits 0 to 2, on the datasets where the gate improves on fixed mixing; Figure~\ref{fig:app_alpha} shows all five.}
\label{fig:alpha_degree}
\end{figure}

\paragraph{The role of degree.}
A degree-dependent self weight recovers the gain of the learned policy. A gate that reads only the degree performs as well as \textsc{TaskGate} on every dataset, whereas fixed mixing with the degree as an extra input feature does not, so the benefit comes from mixing that adapts to the degree rather than from the degree as information; Appendix~\ref{app:random_policy} also compares fixed degree rules and normalizations of the neighborhood channel. The policy uses the degree differently across datasets (Figure~\ref{fig:alpha_degree}; Appendix~\ref{app:where}): on Questions and Tolokers, high-degree nodes get a lower self weight, whereas on Roman-empire, where the gain is largest, they keep more of their own channel than low-degree nodes in the early and middle layers and much less in the last four.

\subsection{Training the gate on the distinctness potential}
\label{sec:objective}

\paragraph{Potential ascent.}
At the selected \textsc{TaskGate} configuration, we compare task training with ascent and descent of the distinctness potential and with the frozen random policy (Appendix~\ref{app:random_policy}), and we vary the gate learning rate, which moves the actions toward the endpoints (Appendix~\ref{app:dose}). Ascent matches the random policy where the gate helps and stays at fixed mixing on Amazon-ratings and Minesweeper, where the random policy loses accuracy; task training is ahead of both on every dataset. Ascent is significantly better than descent on Roman-empire and Tolokers (Appendix~\ref{app:stats}).

\paragraph{Distance from equilibrium.}
A shared policy trained on the potential is not guaranteed to reach a Nash equilibrium (Proposition~\ref{prop:amortized}), so we measure each node's unilateral regret in trained networks with frozen channels; all regrets vanish exactly at an equilibrium, and their sum is the NashConv distance \citep{lanctot2017unifiedgametheoreticapproachmultiagent}. Since each utility is convex in the node's own action, extreme actions lower regret even at the wrong endpoint, so we compare with the actions permuted among the nodes (Appendix~\ref{app:probe}). As Proposition~\ref{prop:endpoint} predicts, larger gate learning rates push the potential-trained actions toward the endpoints, where their regret falls clearly below that of their permutation while accuracy falls below fixed mixing on most datasets. At the small rates that validation selects on most datasets, the two regrets are close, and the task-trained policy has about the same regret as uniform mixing.

\paragraph{Cost.}
The policy adds six parameters per layer to the two transforms of a GraphSAGE layer; a training step takes about 1.2 times as long as with fixed mixing, and twice as long for \textsc{GameGate} (Appendix~\ref{app:cost}).
\section{Conclusion}
\label{sec:conclusion}

We formulated task-based and distinctness-based self-neighborhood mixing as two potential games over the same players and actions, and introduced \method{}, which amortizes the actions through a shared six-parameter policy per layer: a lightweight inductive rule that applies directly to unseen nodes and graphs, without graph-specific best-response dynamics.

In our experiments, this small policy improves on fixed mixing on both benchmarks. On the heterophily suite, varying the mixing across nodes helps on several datasets and learning the policy adds to it; a degree-dependent self weight recovers this gain, which makes node degree an informative signal for controlling local propagation. On ECHO, the clearest gain is on eccentricity. The potential-game view characterizes the layer-local incentives exactly, and its prediction of hard endpoint equilibria is borne out when the gate is trained on the potential, while the best accuracy comes from task-trained actions. Shared low-capacity policies trained on the task loss are a practical way to turn node-wise channel decisions into inductive layers.

\paragraph{Limitations.}
Each node of \method{} chooses one scalar mixture of its self and strict-neighborhood channels; finer-grained choices, such as neighbor selection, propagation range, or rewiring, are beyond its scope. Its benefits vary across tasks, and specialized methods remain ahead on several ECHO tasks. The distinctness potential measures variation in representation space rather than task utility, and its equilibrium analysis assumes fixed channels and independent actions, whereas the learned policy shares parameters across nodes and trains jointly with the channels; extending the analysis to such policies is a natural next step.
\subsection*{AI use statement}

In this work, we used generative AI tools for feedback on the experimental design and the control experiments, and for assistance in implementing and verifying selected experimental code (porting methods, writing smoke tests and scripts that check reported values, and troubleshooting runs), in interpreting results, and in writing and checking proofs. We have not used generative AI tools for developing the theoretical model, formulating hypotheses, generating synthetic data, translating content, or cleaning or reformatting datasets, and qualitative and thematic data analysis are not applicable to this work. Additionally, we used generative AI tools for editing software code and improving the readability of the manuscript. We have reviewed all AI-assisted work. The authors inspected and tested all AI-assisted code, checked the verification scripts against the experimental logs and all AI-assisted proofs line by line, and revised all AI-edited text. We take responsibility for the final content of this work, including text, claims, or artifacts produced with the aid of generative AI.

\bibliography{references}

\clearpage
\appendix

\section*{Supplementary Material}
\paragraph{Overview.}
Appendix~\ref{app:setup} gives the full setup and Appendix~\ref{app:stats} the statistical tests. The other appendices follow the main text.
\begin{itemize}
\item Section~\ref{sec:game}: normalization of the energy, amortized play, finite improvement, and minibatching (Appendix~\ref{app:theory}).
\item Section~\ref{sec:benchmarks}: the backbone recipe added to the channel mixers, the published settings of \citet{luo2024classic}, and the ECHO protocol and ablations (Appendix~\ref{app:hetero}).
\item Section~\ref{sec:free}: the mixing ladder, the frozen random policy with the degree and normalization controls, where the gate acts, and policy transfer (Appendix~\ref{app:policy}).
\item Section~\ref{sec:objective}: the gate learning rate and the direction of the potential, scarce labels, and the unilateral regret of the trained gates (Appendix~\ref{app:objective}).
\item The cost paragraph of Section~\ref{sec:experiments}: parameters and time per layer (Appendix~\ref{app:cost}).
\end{itemize}

\FloatBarrier
\section{Experimental Setup}
\label{app:setup}

We use three experimental protocols. The main heterophily comparison selects one configuration per model and dataset by mean validation performance within that model's evaluated search space. Heterophily diagnostics either reuse a selected configuration or vary one architectural or optimization component and are not included in the main ranking. ECHO experiments follow the benchmark data, metrics, and evaluation protocol of \citet{miglior2026can}, with one stated configuration for every model (Appendix~\ref{app:echo}).

\paragraph{Data and preprocessing.}

The datasets are obtained from their public releases and require no manual preprocessing. We use the ten official splits of the heterophily suite of \citet{platonov2024criticallookevaluationgnns}, the official ECHO loaders and splits of \citet{miglior2026can}, and the synthetic graph-property benchmark used in A-DGN \citep{gravina2023antisymmetric} to train the policies evaluated in the barbell probe. The heterophily graphs are symmetrized by the transductive harness. The \method{} neighborhood channel removes self-loop edges, so $\mathcal{N}'(i)$ remains strict even when the backbone edge index contains self loops. Gate descriptors are standardized over the nodes of each input graph. We apply no additional filtering or transformations beyond those defined by the corresponding benchmark loaders and evaluation protocols. The transductive harness and classic baselines are based on the release of \citet{luo2024classic}.

\paragraph{Datasets.}
We use the five heterophily datasets of \citet{platonov2024criticallookevaluationgnns}: Roman-empire, Amazon-ratings, Minesweeper, Questions, and Tolokers. Table~\ref{tab:app_data_hetero} reports their statistics. ECHO contains three synthetic graph tasks and two molecular tasks; their train, validation, and test sizes are reported in Table~\ref{tab:app_data_echo}. All official splits are used unchanged.

\begin{table}[tbp]
\centering\small
\setlength{\tabcolsep}{3pt}
\begin{tabular}{@{}lrrrrr@{}}
\toprule
Dataset & Nodes & Edges & Feat. & Cls. & Hom. \\
\midrule
Roman-empire   & 22{,}662 & 32{,}927  & 300 & 18 & 0.05 \\
Amazon-ratings & 24{,}492 & 93{,}050  & 300 & 5  & 0.38 \\
Minesweeper    & 10{,}000 & 39{,}402  & 7   & 2  & 0.68 \\
Tolokers       & 11{,}758 & 519{,}000 & 10  & 2  & 0.59 \\
Questions      & 48{,}921 & 153{,}540 & 301 & 2  & 0.84 \\
\bottomrule
\end{tabular}
\caption{Heterophily suite of \citet{platonov2024criticallookevaluationgnns}, used transductively with
the ten official splits. Hom.\ is edge homophily (fraction of edges joining
same-label nodes). Roman-empire and Amazon-ratings are scored by accuracy, the
other three by ROC-AUC.}
\label{tab:app_data_hetero}
\end{table}

\begin{table}[!tbp]
\centering\small
\setlength{\tabcolsep}{4pt}
\begin{tabular}{@{}llrrr@{}}
\toprule
Task & Target & Train & Valid & Test \\
\midrule
\grp{5}{ECHO-Synth (graphs)}\\
SSSP     & node  & 5{,}760 & 2{,}160 & 2{,}160 \\
ECC      & node  & 5{,}760 & 2{,}160 & 2{,}160 \\
Diameter & graph & 5{,}760 & 2{,}160 & 2{,}160 \\
\midrule
\grp{5}{ECHO-Chem (molecules)}\\
Charge & node  & 153{,}330 & 8{,}518 & 8{,}519 \\
Energy & graph & 157{,}222 & 19{,}652 & 19{,}654 \\
\bottomrule
\end{tabular}
\caption{ECHO long-range benchmark \citep{miglior2026can}: counts of the
train, validation, and test graphs shipped with each task, and whether the
regression target is per node or per graph.}
\label{tab:app_data_echo}
\end{table}

\paragraph{Evaluation metrics.}
Roman-empire and Amazon-ratings are evaluated with accuracy. Minesweeper, Questions, and Tolokers use ROC-AUC, following \citet{platonov2024criticallookevaluationgnns}. The latter three datasets are binary and class-imbalanced; for example, a constant majority predictor reaches $80.0$ accuracy but $50.0$ ROC-AUC on Minesweeper.

ECHO training minimizes the benchmark $\log_{10}$ MSE objective on the target representation provided by the loader. Synthetic-task MAE is converted back to the benchmark scale by multiplying by $40$. Molecular energy is evaluated in the original target space after inverting its logarithmic representation, while charge is displayed in $10^{-3}$ units. Test MAE is reported at the epoch with the lowest validation loss.

\paragraph{Training budget and optimization: heterophily.}

All heterophily models receive 500 backbone-training epochs with Adam, weight decay $5\!\times\!10^{-4}$, and full-batch updates. The neighborhood channel of \method{} uses mean aggregation, and every gate policy starts from zero weights and bias, so that all actions equal $1/2$ at initialization. Appendix~\ref{app:schedule} gives the schedule of \textsc{GameGate}.

\subsection{Search spaces and selected configurations}

Tables~\ref{tab:app_search} and~\ref{tab:app_selected} list the evaluated search spaces and the selected configurations. The grids are model-specific because the architectures expose different hyperparameters. For each model and dataset, we select the configuration with the highest mean validation score within that model's grid; test performance is not used for selection. Because the official splits overlap, the validation nodes of one split include test nodes of others, so selecting by the mean over splits lets those labels inform the choice; this is common practice on this suite and affects every model alike. Hyperparameters not listed are fixed, and the only baseline-specific switch in the search is the separate channel projection of CHAT-GNN. On ECHO, every model we run uses one configuration, with twenty layers, width 64, and learning rate $10^{-3}$ (Appendix~\ref{app:echo}).

\begin{table}[!htbp]
\centering
\caption{Search spaces for the heterophily comparison, counted from the run records. Every configuration uses Adam, weight decay $5\cdot10^{-4}$, and the ten official splits; the main comparison trains for at most 500 epochs. Configs is the number of configurations evaluated per dataset. \textsc{GameGate} then selects its gate learning rate from $\{10^{-4},10^{-3},10^{-2},10^{-1}\}$ at its selected configuration, and on Minesweeper, \textsc{Fixed} has one additional candidate at depth 7. The five \method{} variants share one grid, on which the classic backbones also run, with the same residual connections and batch normalization. The diagnostics add these two components of \citet{luo2024classic} (recipe) to the channel mixers, remove them from \textsc{TaskGate}, or train with the longer published budgets.}
{\footnotesize
\setlength{\tabcolsep}{2.8pt}
\begin{tabular}{@{}lcccccl@{}}
\toprule
Family & Width & Depth & Learning rate & Dropout & Configs & Notes \\
\midrule
\grp{7}{Main comparison}\\
\method{} variants & $\{256,512\}$ & $10$ & $\{10^{-3},5{\cdot}10^{-3}\}$ & $\{0.2,0.5\}$ & 8 & shared grid \\
FAGCN & $\{256,512\}$ & $\{2,4,7,8\}$ & $\{10^{-3},5{\cdot}10^{-3},10^{-2}\}$ & $\{0.3,0.5\}$ & 48 & $\varepsilon=0.2$ \\
GPR-GNN & $\{256,512\}$ & $2$ & $\{10^{-3},5{\cdot}10^{-3},10^{-2}\}$ & $\{0.2,0.3,0.5\}$ & 18 & $K=10$, $\alpha=0.1$ \\
ACM-GCN & $256$ & $\{2,4,7,10\}$ & $\{10^{-3},5{\cdot}10^{-3},10^{-2}\}$ & $0.5$ & 12 &  \\
H2GCN & $256$ & $2$ & $\{5{\cdot}10^{-3},10^{-2}\}$ & $\{0,0.5\}$ & 4 & 2 rounds \\
CHAT-GNN & $256$ & $\{2,5,10\}$ & $\{10^{-3},5{\cdot}10^{-3},10^{-2}\}$ & $\{0,0.5\}$ & 16--21 & projection on/off \\
GCNH & $256$ & $\{2,4,8,16\}$ & $\{10^{-3},5{\cdot}10^{-3}\}$ & $\{0.2,0.5\}$ & 16 &  \\
GATE & $\{64,256\}$ & $\{2,5,10\}$ & $\{10^{-3},5{\cdot}10^{-3},10^{-2}\}$ & $\{0,0.5\}$ & 32 &  \\
GCN, SAGE, GAT & $\{256,512\}$ & $10$ & $\{10^{-3},5{\cdot}10^{-3}\}$ & $\{0.2,0.5\}$ & 24 & grid of \method{} \\
\midrule
\grp{7}{Diagnostics}\\
FAGCN, ACM-GCN, GATE & $256$ & $\{4,10\}$ & $\{5{\cdot}10^{-3},10^{-2}\}$ & $0.5$ & 4 & recipe added \\
GCNH & $256$ & $\{2,4,8,16\}$ & $\{10^{-3},5{\cdot}10^{-3}\}$ & $\{0.2,0.5\}$ & 16 & recipe added \\
\textsc{TaskGate} & $256$ & $\{2,4,10\}$ & $\{10^{-3},5{\cdot}10^{-3}\}$ & $\{0.2,0.5\}$ & 12 & recipe removed \\
\textsc{TaskGate} & $\{256,512\}$ & $\{4,10,16\}$ & $\{3{\cdot}10^{-5},10^{-3}\}$ & $\{0.2,0.3,0.5\}$ & 1--2 & longer budget \\
GCN, SAGE, GAT & $\{64,256,512\}$ & $3$--$15$ & $\{3{\cdot}10^{-5},10^{-3},10^{-2}\}$ & $\{0.2,0.3,0.5\}$ & 3 & published settings \\
\bottomrule
\end{tabular}

}
\label{tab:app_search}
\end{table}

\begin{table}[!htbp]
\centering
\caption{Validation-selected configurations for the heterophily comparison, as width/depth/learning rate/dropout; the best-classic row also names the backbone. The rows below \textsc{GameGate}, CHAT-GNN, and the best classic give the selected gate learning rate, separate-projection setting, and backbone.}
{\footnotesize
\setlength{\tabcolsep}{3pt}
\begin{tabular}{@{}llllll@{}}
\toprule
Model & Roman-empire & Amazon-ratings & Minesweeper & Questions & Tolokers \\
\midrule
\grp{6}{\method{} and the policy ladder}\\
\textsc{TaskGate} & 256/10/0.005/0.2 & 512/10/0.001/0.2 & 512/10/0.001/0.5 & 256/10/0.005/0.5 & 256/10/0.001/0.5 \\
\textsc{GameGate} & 256/10/0.005/0.2 & 512/10/0.001/0.2 & 256/10/0.005/0.2 & 256/10/0.005/0.5 & 256/10/0.005/0.5 \\
\quad gate learning rate & $10^{-3}$ & $10^{-4}$ & $10^{-4}$ & $10^{-2}$ & $10^{-3}$ \\
\textsc{Task-MLP} & 256/10/0.005/0.2 & 512/10/0.001/0.2 & 256/10/0.005/0.5 & 256/10/0.005/0.5 & 256/10/0.001/0.5 \\
\textsc{Fixed} & 256/10/0.005/0.2 & 512/10/0.001/0.2 & 256/10/0.005/0.5 & 256/10/0.001/0.5 & 256/10/0.005/0.5 \\
\textsc{Layer} & 256/10/0.005/0.2 & 512/10/0.001/0.2 & 256/10/0.005/0.5 & 256/10/0.005/0.5 & 256/10/0.005/0.5 \\
\midrule
\grp{6}{Channel mixers}\\
FAGCN & 512/7/0.01/0.3 & 512/2/0.005/0.3 & 256/8/0.01/0.3 & 512/2/0.005/0.3 & 512/2/0.005/0.3 \\
GPR-GNN & 256/2/0.01/0.3 & 256/2/0.01/0.2 & 512/2/0.01/0.2 & 256/2/0.01/0.2 & 256/2/0.01/0.2 \\
ACM-GCN & 256/4/0.005/0.5 & 256/2/0.01/0.5 & 256/2/0.005/0.5 & 256/4/0.005/0.5 & 256/2/0.005/0.5 \\
H2GCN & 256/2/0.01/0.0 & 256/2/0.005/0.0 & 256/2/0.01/0.0 & 256/2/0.01/0.0 & 256/2/0.005/0.0 \\
CHAT-GNN & 256/10/0.005/0.5 & 256/5/0.005/0.5 & 256/5/0.001/0.5 & 256/5/0.005/0.5 & 256/5/0.01/0.0 \\
\quad projection & off & off & on & off & off \\
GCNH & 256/2/0.005/0.5 & 256/2/0.005/0.2 & 256/4/0.005/0.2 & 256/4/0.005/0.5 & 256/4/0.005/0.2 \\
GATE & 256/2/0.01/0.5 & 256/2/0.01/0.5 & 256/5/0.01/0.0 & 256/5/0.005/0.5 & 256/5/0.005/0.0 \\
\midrule
\grp{6}{Classic backbones}\\
best classic & 512/10/0.005/0.5 & 512/10/0.001/0.5 & 256/10/0.005/0.5 & 256/10/0.001/0.5 & 256/10/0.001/0.5 \\
\quad backbone & GCN & SAGE & GCN & GCN & GCN \\
\bottomrule
\end{tabular}

}
\label{tab:app_selected}
\end{table}

\subsection{Training details}

\paragraph{GameGate schedule.}
\label{app:schedule}
The gate policies of \textsc{GameGate} have their own Adam optimizer with weight decay $10^{-4}$. On the heterophily datasets, we first select the configuration of \textsc{GameGate} with gate learning rate $10^{-4}$ and then select the gate learning rate from $\{10^{-4},10^{-3},10^{-2},10^{-1}\}$ at that configuration; on ECHO, it is fixed at $10^{-4}$. The task-only warm-up of Algorithm~\ref{alg:gamegate} lasts 50 epochs on the heterophily datasets, one tenth of their 500-epoch budget, and also 50 epochs on ECHO-Synth, whose runs stop early within at most 1000 epochs. On ECHO-Chem, it lasts 10 epochs, about 6000 mini-batch updates: one tenth of the 100-epoch budget of charge and one thirtieth of the 300-epoch budget of energy. Each round is played on the full graph on the heterophily datasets and on one mini-batch on ECHO. Every reported run uses $n_{\mathrm{game}}=n_{\mathrm{task}}=1$, so a round takes the place of one training step of the other models, which on the full-batch heterophily graphs is one epoch; \textsc{GameGate} therefore performs as many task updates as they do.

Table~\ref{tab:app_schedule} varies both choices at the selected configuration of each heterophily dataset. The warm-up is set to 0, 25, or 100 epochs, and the updates per round to $n_{\mathrm{game}}=n_{\mathrm{task}}=n\in\{5,25\}$, keeping the totals of gate and task updates of the reported runs, so the number of rounds shrinks by a factor $n$ and validation is checked once per round. Neither choice matters much within the tested range: every variant stays within about one standard deviation across splits of the reported runs, with differences of at most 0.6 points on four datasets and 1.0 point on Questions.

\begin{table}[!htbp]
\centering
\caption{Sensitivity of \textsc{GameGate} to its schedule at the selected configuration of each heterophily dataset: mean $\pm$ standard deviation over the ten splits. The first row gives the reported runs.}
\label{tab:app_schedule}
\begin{tabular}{lccccc}
\toprule
Schedule & Roman-emp. & Amazon-rat. & Minesw. & Questions & Tolokers \\
\midrule
as reported (warm-up 50, $n=1$) & 88.4$\pm$0.8 & 53.9$\pm$0.5 & 95.4$\pm$0.7 & 75.6$\pm$1.5 & 84.6$\pm$0.7 \\
\midrule
\grp{6}{Warm-up epochs before the first gate update}\\
warm-up 0 & 88.5$\pm$0.6 & 54.0$\pm$0.6 & 95.1$\pm$0.4 & 75.0$\pm$1.4 & 84.6$\pm$0.6 \\
warm-up 25 & 88.2$\pm$0.8 & 53.9$\pm$0.6 & 95.4$\pm$0.6 & 75.9$\pm$1.2 & 84.4$\pm$0.7 \\
warm-up 100 & 87.8$\pm$0.5 & 54.0$\pm$0.6 & 95.3$\pm$0.5 & 74.6$\pm$1.2 & 84.5$\pm$0.7 \\
\midrule
\grp{6}{Gate updates per round}\\
$n=5$ & 88.3$\pm$0.5 & 53.4$\pm$0.5 & 95.0$\pm$0.7 & 75.0$\pm$1.5 & 84.3$\pm$0.7 \\
$n=25$ & 88.2$\pm$0.6 & 53.5$\pm$0.5 & 95.2$\pm$0.7 & 75.1$\pm$0.9 & 84.2$\pm$0.8 \\
\bottomrule
\end{tabular}

\end{table}

\paragraph{Free per-node actions.}
\label{app:free}
In Section~\ref{sec:free}, every node has its own action at every layer: the first of two softmax weights over a pair of logits, which start equal, so that the action starts at $1/2$. Under the task loss, the actions are either trained with the optimizer of the remaining parameters or given their own Adam group with learning rate in $\{10^{-5},10^{-4},10^{-3},10^{-2}\}$ and weight decay in $\{0,10^{-4}\}$, and validation selects among these nine settings. Under the distinctness potential, the free actions and the shared policy are both selected over the learning rates $10^{-4}$ and $10^{-3}$; on Questions, where \textsc{GameGate} selects $10^{-2}$, the shared-policy entry of Table~\ref{tab:free} therefore differs slightly from Table~\ref{tab:hetero}.

\subsection{Reproducibility}

\paragraph{Seeds and sources of randomness.}

Each heterophily result is a mean over the ten official splits, with one training run per split, so the reported standard deviation reflects both the splits and the training randomness; each ECHO result is a mean over five seeds. Different experiments can draw different random numbers even when they rerun the same model at the same configuration, so its score can differ slightly between tables.

\paragraph{Computing infrastructure.}
Experiments ran on NVIDIA A100-SXM4 40\,GB GPUs in a two-socket AMD EPYC 7742 server with 1\,TB of memory, under Ubuntu~22.04 with Python~3.9.5, PyTorch~2.8.0, CUDA~12.8, and PyTorch Geometric~2.6.1. Each run uses one GPU.


\FloatBarrier
\section{Additional Theoretical Results}
\label{app:theory}

This section clarifies the normalization of the energy and the relation between the games of Section~\ref{sec:game} and the shared action policy used by \method{}, and records two consequences of the potential-game formulation.

\paragraph{Normalization of the energy.}
Eq.~\eqref{eq:dirichlet} divides by $|V|$, although only the nodes of $V_+$ contribute terms; dividing by $|V_+|$ instead would rescale the energy and every utility by the same positive constant and change no best response or equilibrium. Its weights $1/d_i$ act on the unnormalized differences $\mathbf{r}_i^\ell-\mathbf{r}_j^\ell$, so $\mathcal{E}^\ell$ is not the Dirichlet energy of the symmetrically normalized Laplacian used to study oversmoothing \citep{zhou2021dirichlet}, which compares degree-rescaled representations.

\paragraph{Amortized play.}
The equilibrium analysis of Section~\ref{sec:game} treats the node actions as independent variables, whereas \method{} generates them through a shared policy. To characterize what carries over to this amortized setting, let $\Psi^\ell$ denote either potential at layer $\ell$, with corresponding utilities $u_i^\ell$. We fix the channel values as in Theorem~\ref{thm:potential}; for the task potential, we likewise fix all network quantities other than the gate parameters. We also fix the descriptors $\mathbf{q}_i^\ell$, consistently with their detached use in Eq.~\eqref{eq:policy}.

Define the marginal payoff of player $i$ as
\[
g_i^\ell
=
\frac{\partial u_i^\ell}{\partial \alpha_i^\ell},
\]
and let
\[
F^\ell(\boldsymbol{\phi}^\ell)
=
\Psi^\ell
\left(
\boldsymbol{\alpha}^\ell(\boldsymbol{\phi}^\ell)
\right)
\]
be the potential evaluated at the action profile generated by the shared policy. We write $[\mathbf{q}_i^\ell;1]\in\mathbb{R}^6$ for the descriptor augmented by a constant coordinate for the bias.

\begin{proposition}[Amortized play]
\label{prop:amortized}
For the shared policy in Eq.~\eqref{eq:policy}:

\begin{enumerate}
    \item
    The gradient of the potential with respect to the shared policy
    parameters is
    \[
    \nabla_{\boldsymbol{\phi}^\ell}F^\ell
    =
    \sum_{i\in V_+}
    g_i^\ell\,
    \alpha_i^\ell(1-\alpha_i^\ell)
    [\mathbf{q}_i^\ell;1].
    \]

    \item
    $\boldsymbol{\phi}^\ell$ is a stationary point of $F^\ell$ if
    and only if
    \[
    \sum_{i\in V_+}
    g_i^\ell\,\delta\alpha_i^\ell
    =
    0
    \]
    for every first-order change
    $\delta\boldsymbol{\alpha}^\ell$
    of the action profile that can be induced by an infinitesimal
    change of the shared policy parameters.

    \item
    If each player instead has its own finite logit
    $\theta_i^\ell$, with
    $\alpha_i^\ell=\sigma(\theta_i^\ell)$, then stationarity is
    equivalent to
    \[
    g_i^\ell=0
    \qquad
    \text{for every } i\in V_+.
    \]
\end{enumerate}
\end{proposition}

\noindent\emph{Proof.}
Because both games are exact potential games,
\[
u_i^\ell(a,\boldsymbol{\alpha}_{-i}^\ell)
-
\Psi^\ell(a,\boldsymbol{\alpha}_{-i}^\ell)
\]
does not depend on $a$. Hence, wherever the potential is differentiable,
\[
\frac{\partial \Psi^\ell}{\partial\alpha_i^\ell}
=
\frac{\partial u_i^\ell}{\partial\alpha_i^\ell}
=
g_i^\ell.
\]
With the descriptors fixed, Eq.~\eqref{eq:policy} gives
\[
\nabla_{\boldsymbol{\phi}^\ell}\alpha_i^\ell
=
\alpha_i^\ell(1-\alpha_i^\ell)
[\mathbf{q}_i^\ell;1].
\]
Applying the chain rule yields part~(i).

For an infinitesimal parameter change $\delta\boldsymbol{\phi}^\ell$, the induced first-order action change satisfies
\[
\delta F^\ell
=
\sum_{i\in V_+}
g_i^\ell\,\delta\alpha_i^\ell.
\]
Therefore $\nabla_{\boldsymbol{\phi}^\ell}F^\ell=\mathbf{0}$ if and only if this quantity vanishes for every first-order action change accessible through the shared policy, proving part~(ii).

Finally, with one finite logit per player,
\[
\frac{\partial F^\ell}{\partial\theta_i^\ell}
=
g_i^\ell\,
\alpha_i^\ell(1-\alpha_i^\ell).
\]
For finite $\theta_i^\ell$, we have
$0<\alpha_i^\ell<1$, and hence
$\alpha_i^\ell(1-\alpha_i^\ell)>0$.
Thus
$\partial F^\ell/\partial\theta_i^\ell=0$
if and only if $g_i^\ell=0$, proving part~(iii).
\hfill$\square$

Proposition~\ref{prop:amortized} separates the game from its amortized implementation. The game determines each player's local incentive $g_i^\ell$, while the shared policy determines which joint changes of the action profile are available during optimization. Consequently, stationarity of the shared policy imposes the players' first-order conditions only along directions representable by that policy and does not, in general, imply individual optimality or a Nash equilibrium. With free per-node logits, by contrast, every action has its own first-order direction, so stationarity requires $g_i^\ell=0$ for every player.

In the distinctness game, such interior stationary points of free actions are not local maxima when the corresponding node is nondegenerate. By Proposition~\ref{prop:endpoint}, the player's payoff is then strictly convex in its own action; an interior point with $g_i^\ell=0$ is therefore its minimum along that unilateral direction, and moving the action in either direction increases both the player's payoff and the potential. Section~\ref{sec:free} compares these free actions with the shared policy empirically.

\paragraph{Finite improvement in the binary game.}
Consider the distinctness game with the channel values fixed as in Theorem~\ref{thm:potential}, and restrict every player to the binary action set
\[
\mathcal{A}_i=\{0,1\}.
\]
The resulting game has finitely many action profiles. By Theorem~\ref{thm:potential}, every unilateral deviation that strictly increases the deviating player's payoff increases $\mathcal{E}^\ell$ by exactly the same amount. Therefore, along any sequence of strict better responses, the potential increases strictly at every step. Since the action space is finite, every such sequence terminates after finitely many updates at a pure Nash equilibrium.

At nondegenerate nodes, this restriction does not remove any best responses: Proposition~\ref{prop:endpoint} already places every best response in $\{0,1\}$. Thus the binary game preserves the best-response structure of the continuous game at those nodes while also satisfying the finite improvement property.

\paragraph{Task game under minibatching.}
On ECHO, the task loss is evaluated on minibatches of training graphs. For any fixed minibatch, consider the union of its non-isolated nodes as the players and assign every player the common utility
\[
u_{i,\mathrm{task}}^\ell
=
-\mathcal{L}_{\mathrm{task}}.
\]
The resulting game remains an identical-interest game: every unilateral action change produces the same change in the player's utility and in the shared payoff $-\mathcal{L}_{\mathrm{task}}$. Therefore, for each fixed minibatch, the task objective remains an exact potential.

\FloatBarrier
\section{Statistical Tests}
\label{app:stats}

We compare two models on a dataset using their scores on the ten official splits. The observations are paired because both models are trained and evaluated on the same split. We therefore apply a two-sided Wilcoxon signed-rank test to the ten paired differences. With ten nonzero paired differences, the smallest attainable two-sided exact $p$-value is $2/2^{10}\approx0.002$, obtained when all signed differences point in the same direction.

All scores are taken from the ten-split runs underlying Table~\ref{tab:hetero}, the mixing ladder (Appendix~\ref{app:ladder}), and the gate learning rate sweep (Appendix~\ref{app:dose}). The original search of CHAT-GNN did not retain per-split scores on four datasets, so its validation-selected configurations were rerun on all ten splits; these reruns are also those reported in Table~\ref{tab:hetero}.

We organize the paired tests into three families and apply Holm's step-down correction separately within each family. The main family contains twenty comparisons: \textsc{TaskGate} against fixed mixing, \textsc{GameGate} with the gate learning rate selected by validation, as in Table~\ref{tab:hetero}, the best classic backbone, and CHAT-GNN on each of the five datasets. CHAT-GNN is the strongest channel mixer in the main comparison on four of the five datasets.

Two additional families contain ten tests each. The first studies policy capacity and granularity by comparing \textsc{TaskGate} with one coefficient per layer (\textsc{Layer}) and with the larger \textsc{Task-MLP} policy. The second studies the distinctness objective by comparing \textsc{GameGate} with fixed mixing and potential ascent with potential descent. With only ten paired observations, the discrete signed-rank test together with Holm correction makes statistical significance require highly consistent paired differences across splits.

In the main family, \textsc{TaskGate} is significantly better than fixed mixing on Roman-empire and Tolokers, and than \textsc{GameGate} on Roman-empire. Against the best classic backbone, it is significantly better on Roman-empire and significantly worse on Minesweeper. It is significantly better than CHAT-GNN on every dataset except Questions. Four additional comparisons have raw $p<0.05$ but do not survive Holm correction: \textsc{TaskGate} against fixed mixing on Questions ($p_{\mathrm{raw}}=0.037$, $p_{\mathrm{Holm}}=0.37$), against the best classic backbone on Tolokers ($p_{\mathrm{raw}}=0.006$, $p_{\mathrm{Holm}}=0.06$), and against \textsc{GameGate} on Questions and Tolokers ($p_{\mathrm{raw}}=0.037$ and $0.049$, $p_{\mathrm{Holm}}=0.37$ and $0.39$).

In the policy family, \textsc{TaskGate} is significantly better than the per-layer coefficient on Roman-empire, Questions, and Tolokers, and significantly worse on Minesweeper; no comparison with \textsc{Task-MLP} is significant after correction. In the distinctness-objective family, \textsc{GameGate} improves significantly over fixed mixing on Roman-empire, while potential ascent is significantly better than descent on Roman-empire and Tolokers.

The ten official splits of a dataset share one graph and overlap in their nodes, so the paired differences are not independent, and with one training run per split, the variation of the seeds is confounded with that of the splits; both make the tests somewhat anti-conservative. Appendix~\ref{app:random_policy} therefore repeats the matched comparisons with five seeds per split and tests them with the corrected resampled $t$-test of \citet{nadeau2003inference}, which allows for the overlap.

\begin{table}[!htbp]
\centering
\caption{Paired comparisons over the ten official splits, grouped by test family. Each entry reports the mean paired score difference, first model minus second, in accuracy points on Roman-empire and Amazon-ratings and ROC-AUC points on the remaining datasets; positive values therefore favor the first model. Statistical significance is assessed separately using two-sided Wilcoxon signed-rank tests with Holm correction within each block. $^{**}$ marks comparisons that remain significant at $0.05$ after correction, and $^{\dagger}$ marks raw $p<0.05$ that do not. On Amazon-ratings, five comparisons have one or two zero paired differences and use the normal approximation of the test.}
\label{tab:app_signif_families}
\begin{tabular}{lccccc}
\toprule
Difference & Roman-emp. & Amazon-rat. & Minesw. & Questions & Tolokers \\
\midrule
\grp{6}{Main comparisons (Table~\ref{tab:hetero}), twenty tests}\\
\textsc{TaskGate} $-$ \textsc{Fixed} & $+1.93^{**}$ & $+0.20$ & $-0.08$ & $+1.79^{\dagger}$ & $+0.99^{**}$ \\
\textsc{TaskGate} $-$ \textsc{GameGate} & $+0.81^{**}$ & $+0.07$ & $-0.01$ & $+1.38^{\dagger}$ & $+0.48^{\dagger}$ \\
\textsc{TaskGate} $-$ best classic & $+0.77^{**}$ & $-0.18$ & $-1.06^{**}$ & $-0.05$ & $-0.51^{\dagger}$ \\
\textsc{TaskGate} $-$ CHAT-GNN & $+0.72^{**}$ & $+2.51^{**}$ & $+2.65^{**}$ & $+0.26$ & $+1.50^{**}$ \\
\midrule
\grp{6}{Gate granularity and capacity, ten tests}\\
\textsc{TaskGate} $-$ \textsc{Layer} & $+1.68^{**}$ & $-0.29$ & $-0.82^{**}$ & $+3.25^{**}$ & $+1.00^{**}$ \\
\textsc{TaskGate} $-$ \textsc{Task-MLP} & $-0.04$ & $+0.17$ & $+0.02$ & $+0.53$ & $-0.01$ \\
\midrule
\grp{6}{Distinctness potential, ten tests}\\
\textsc{GameGate} $-$ \textsc{Fixed} & $+1.12^{**}$ & $+0.14$ & $-0.06$ & $+0.41$ & $+0.52^{\dagger}$ \\
ascent $-$ descent & $+1.09^{**}$ & $-0.10$ & $+0.07$ & $+0.07$ & $+0.40^{**}$ \\
\bottomrule
\end{tabular}

\end{table}


\FloatBarrier
\section{Heterophily and Long-Range Benchmarks}
\label{app:hetero}

This section complements Section~\ref{sec:benchmarks}. Unless stated otherwise, heterophily entries here and in Appendices~\ref{app:policy} and~\ref{app:objective} are means and standard deviations over the ten official splits, and each variant is selected by validation within the search space of Table~\ref{tab:app_search}.

\subsection{Backbone recipe and published settings}

\paragraph{Backbone recipe.}
The channel mixers of Table~\ref{tab:hetero} run in their published form, whereas \method{} and the classic backbones use the residual connections and batch normalization of \citet{luo2024classic}. All channel-mixer scores are our reruns under this protocol and can differ from published scores obtained under other protocols. Table~\ref{tab:app_recipe} adds these two components to FAGCN, ACM-GCN, GCNH, and GATE. They improve all four mixers on every dataset, except GCNH on Tolokers, which changes by 0.1 points. With them, the best mixer on each dataset comes within half a point of \textsc{TaskGate} and exceeds it on Roman-empire, Minesweeper, and Questions, although no single mixer comes that close on all five datasets. Removing the two components from \method{} lowers \textsc{TaskGate} on every dataset, most on Roman-empire and Amazon-ratings. Most of the margin of \method{} over the published mixers thus disappears once the mixers receive the same two components, and Section~\ref{sec:free} measures the gate against fixed mixing on the same backbone. We do not add the recipe to GPR-GNN, H2GCN, or CHAT-GNN, whose propagation or readout already incorporates related components.

\paragraph{Published settings.}
The remaining rows of Table~\ref{tab:app_recipe} leave the matched budget for the longer, model-specific budgets of the published configurations of \citet{luo2024classic}, and they contain two versions of \textsc{TaskGate}. The first keeps its own configuration and only lengthens the budget, which changes it by at most one point; the best classic backbone is ahead of it on all four datasets with published configurations. In the second, the \method{} layer replaces the classic layer in the published configurations, with the gate or with fixed mixing. There, the gate improves on fixed mixing by 1.4 points on Roman-empire and slightly on Questions, and lowers the score on Minesweeper (Table~\ref{tab:app_published_signif}), and \textsc{TaskGate} is significantly ahead of the classic backbone on Roman-empire and significantly behind it on Minesweeper and Questions.

\begin{table}[!htbp]
\centering
\caption{Backbone recipe and published settings: mean $\pm$ standard deviation over the ten splits, with \textsc{TaskGate} from Table~\ref{tab:hetero} in the first row. The recipe is the residual connections and batch normalization of \citet{luo2024classic}. Their published configurations come with longer, model-specific budgets; in them, the \method{} layer replaces the classic layer, with the gate or with fixed mixing, while the last row of that block keeps the \textsc{TaskGate} configuration and only lengthens its budget. Tolokers has no run at the published configurations (--).}
\label{tab:app_recipe}
\begin{tabular}{lccccc}
\toprule
Model & Roman-emp. & Amazon-rat. & Minesw. & Questions & Tolokers \\
\midrule
\textsc{TaskGate} (Table~\ref{tab:hetero}) & 89.2$\pm$0.4 & 54.0$\pm$0.3 & 95.4$\pm$0.4 & 77.0$\pm$2.1 & 85.1$\pm$0.5 \\
\midrule
\grp{6}{Channel mixers with the recipe, 500 epochs}\\
FAGCN & 82.4$\pm$0.8 & 48.2$\pm$0.7 & 90.7$\pm$0.7 & 77.6$\pm$1.2 & 81.7$\pm$1.0 \\
ACM-GCN & 89.4$\pm$0.5 & 48.9$\pm$0.6 & 92.0$\pm$0.5 & 77.5$\pm$1.0 & 81.6$\pm$1.1 \\
GCNH & 89.3$\pm$0.7 & 51.1$\pm$0.6 & 94.5$\pm$0.3 & 77.2$\pm$1.1 & 84.2$\pm$1.0 \\
GATE & 86.1$\pm$0.6 & 53.6$\pm$0.6 & 95.7$\pm$1.0 & 75.1$\pm$1.0 & 85.1$\pm$0.7 \\
\midrule
\grp{6}{\method{} without the recipe, 500 epochs}\\
\textsc{TaskGate} & 80.2$\pm$0.7 & 47.8$\pm$1.1 & 92.2$\pm$0.5 & 76.4$\pm$1.0 & 83.8$\pm$1.2 \\
\midrule
\grp{6}{Published configurations of \citet{luo2024classic}, up to 2500 epochs}\\
best classic & 91.0$\pm$0.4 & 55.2$\pm$0.5 & 97.5$\pm$0.4 & 78.3$\pm$1.3 & -- \\
\textsc{TaskGate} in these configurations & 91.7$\pm$0.4 & 54.9$\pm$0.6 & 96.5$\pm$0.6 & 76.8$\pm$1.1 & -- \\
\textsc{Fixed} in these configurations & 90.3$\pm$0.5 & 55.3$\pm$0.3 & 97.2$\pm$0.4 & 76.6$\pm$1.0 & -- \\
\textsc{TaskGate}, own configuration & 90.2$\pm$0.3 & 54.4$\pm$0.6 & 95.3$\pm$0.4 & 76.5$\pm$1.0 & 85.0$\pm$0.3 \\
\bottomrule
\end{tabular}

\end{table}

\begin{table}[!htbp]
\centering
\caption{Paired tests at the published configurations of \citet{luo2024classic} (Table~\ref{tab:app_recipe}), first model minus second, with the markers of Table~\ref{tab:app_signif_families} and Holm's correction over the eight tests; Tolokers has no run at these configurations (--).}
\label{tab:app_published_signif}
\begin{tabular}{lccccc}
\toprule
Difference & Roman-emp. & Amazon-rat. & Minesw. & Questions & Tolokers \\
\midrule
\grp{6}{Published configurations of \citet{luo2024classic}}\\
\textsc{TaskGate} $-$ \textsc{Fixed} & $+1.44^{**}$ & $-0.36$ & $-0.62^{**}$ & $+0.21^{**}$ & -- \\
\textsc{TaskGate} $-$ best classic & $+0.73^{**}$ & $-0.30$ & $-1.01^{**}$ & $-1.46^{**}$ & -- \\
\bottomrule
\end{tabular}

\end{table}

\FloatBarrier
\subsection{ECHO protocol and ablations}
\label{app:echo}

\paragraph{Protocol.}

We use the readout and training loss of the benchmark \citep{miglior2026can}: a two-layer MLP on the node representations for node-level tasks; mean, max, and sum pooling, concatenated and followed by a two-layer MLP, for graph-level tasks; and the base-10 logarithm of the mean squared error, minimized with Adam, with early stopping on the validation loss. The benchmark authors tune every model on every task with a 100-trial Bayesian search over, among others, depth, width, learning rate, and weight decay; the best results they report, quoted in the last rows of Tables~\ref{tab:echo_synth} and~\ref{tab:echo_chem}, come from that search. We do not tune per model: every model we run uses 20 layers, width 64, ReLU activations, learning rate $10^{-3}$, no weight decay or dropout, and batch size 256, for at most 1000 epochs with patience 50 on ECHO-Synth, 100 epochs with patience 25 on charge, and 300 epochs with patience 50 on energy. \method{}, GCN, GraphSAGE, and GAT use residual connections, batch normalization, and a jumping-knowledge sum over layers \citep{xu2018jk}; the channel mixers are run as published, without these additions, so that each is evaluated in its original form, which for GATE means no residual connections or normalization. On ECHO-Chem, only \method{} uses the bond attributes, through a linear embedding added to each neighborhood message; Table~\ref{tab:app_charge} includes \method{} runs without them. Besides early stopping, validation selects only the aggregation of \method{} (Table~\ref{tab:app_echo_aggr}) and the model of the best classic row; the gate learning rate of \textsc{GameGate} is fixed at $10^{-4}$ (Appendix~\ref{app:schedule}). Under this configuration, our GCN has a lower test error than the GCN of the benchmark's search on four of the five tasks and is close on the fifth, so the shared configuration does not handicap a standard message-passing model. In this configuration, GATE does not fit the training data of charge and eccentricity.
\paragraph{Ablations.}

Sum aggregation lowers the shortest-path error of every variant, and on eccentricity the node-wise gates improve on fixed mixing and on one coefficient per layer under both aggregations (Table~\ref{tab:app_echo_aggr}); diameter under sum aggregation varies strongly across seeds. On charge with the bond attributes, the gate brings no clear gain over fixed mixing with separate transforms; without the bond attributes, both are worse and the gate improves on fixed mixing (Table~\ref{tab:app_charge}).

\begin{table}[!htbp]
\centering
\caption{Gate granularity under each aggregation on ECHO-Synth: test MAE in the units of Table~\ref{tab:echo_synth}, mean $\pm$ standard deviation over five seeds. The main tables select one aggregation per row by validation loss.}
\label{tab:app_echo_aggr}
\begin{tabular}{lccc}
\toprule
Model & SSSP & ECC & Diam. \\
\midrule
\grp{4}{Mean aggregation}\\
\textsc{Fixed} & 2.18$\pm$0.23 & 5.39$\pm$0.09 & 2.56$\pm$0.26 \\
\textsc{Layer} & 2.25$\pm$0.30 & 5.42$\pm$0.09 & 2.79$\pm$0.68 \\
\textsc{TaskGate} & 1.18$\pm$0.13 & 3.49$\pm$0.18 & \snd{1.52$\pm$0.16} \\
\textsc{Task-MLP} & 1.10$\pm$0.32 & \trd{3.38$\pm$0.21} & \best{1.45$\pm$0.13} \\
\textsc{GameGate} & 1.48$\pm$0.85 & 4.80$\pm$0.38 & \trd{1.59$\pm$0.08} \\
\midrule
\grp{4}{Sum aggregation}\\
\textsc{Fixed} & \snd{0.72$\pm$0.07} & 5.15$\pm$0.09 & 2.68$\pm$0.86 \\
\textsc{Layer} & \trd{0.74$\pm$0.18} & 5.21$\pm$0.05 & 2.50$\pm$0.90 \\
\textsc{TaskGate} & \best{0.63$\pm$0.03} & \snd{3.30$\pm$0.15} & 4.16$\pm$1.72 \\
\textsc{Task-MLP} & \best{0.63$\pm$0.05} & \best{3.21$\pm$0.12} & 2.06$\pm$0.38 \\
\textsc{GameGate} & \snd{0.72$\pm$0.15} & 4.35$\pm$0.44 & 3.19$\pm$1.19 \\
\bottomrule
\end{tabular}

\end{table}

\begin{table}[!htbp]
\centering
\caption{Charge (test MAE, $10^{-3}$ units, five seeds): gate granularity, tied transforms, and runs without the bond attributes, under both aggregations; -- marks aggregations that were not run.}
\label{tab:app_charge}
\begin{tabular}{lcc}
\toprule
Model & Mean aggr. & Sum aggr. \\
\midrule
\grp{3}{Separate transforms, bond attributes}\\
\textsc{Fixed} & \trd{5.69$\pm$0.15} & \snd{5.69$\pm$0.09} \\
\textsc{Layer} & 5.74$\pm$0.16 & -- \\
\textsc{TaskGate} & \best{5.57$\pm$0.18} & \best{5.66$\pm$0.06} \\
\midrule
\grp{3}{Tied transforms, bond attributes}\\
\textsc{Fixed} & 5.92$\pm$0.10 & -- \\
\textsc{TaskGate} & \snd{5.61$\pm$0.05} & -- \\
\midrule
\grp{3}{Separate transforms, no bond attributes}\\
\textsc{Fixed} & 7.01$\pm$0.10 & -- \\
\textsc{TaskGate} & 6.57$\pm$0.08 & -- \\
\bottomrule
\end{tabular}

\end{table}


\FloatBarrier
\section{Free versus Amortized Strategies}
\label{app:policy}
This section complements Section~\ref{sec:free}.

\subsection{Mixing ladder}
\label{app:ladder}
Table~\ref{tab:app_ladder} gives the five \method{} variants compared in Section~\ref{sec:free}, on the same backbone and eight-configuration grid; Appendix~\ref{app:stats} tests their differences. One coefficient per layer spans the same functions as fixed mixing, since a constant action can be absorbed into the two transforms, so the differences between \textsc{Layer} and \textsc{Fixed}, from $-1.5$ to $+0.8$ points, come from the parameterization and its optimization rather than from expressivity, and give a scale for the other contrasts of the ladder.

\begin{table}[!htbp]
\centering
\caption{Mixing ladder: the five \method{} variants on the same backbone and eight-configuration grid, grouped by how the gate is trained. Mean $\pm$ standard deviation over the ten splits; bold, underline, and italics mark the three best values per column.}
\label{tab:app_ladder}
\begin{tabular}{lccccc}
\toprule
Model & Roman-emp. & Amazon-rat. & Minesw. & Questions & Tolokers \\
\midrule
\grp{6}{Gate off}\\
\textsc{Fixed} ($\alpha=1/2$) & 87.3$\pm$0.6 & 53.8$\pm$0.5 & \snd{95.4$\pm$0.4} & 75.2$\pm$1.3 & \trd{84.1$\pm$0.6} \\
\midrule
\grp{6}{Trained on the task loss}\\
\textsc{Layer}, one coefficient per layer & 87.5$\pm$0.7 & \best{54.3$\pm$0.6} & \best{96.2$\pm$0.4} & 73.8$\pm$1.5 & \trd{84.1$\pm$0.5} \\
\textsc{TaskGate}, six-parameter policy & \snd{89.2$\pm$0.4} & \snd{54.0$\pm$0.3} & \snd{95.4$\pm$0.4} & \best{77.0$\pm$2.1} & \best{85.1$\pm$0.5} \\
\textsc{Task-MLP}, MLP policy & \best{89.3$\pm$0.5} & 53.8$\pm$0.6 & \trd{95.3$\pm$0.4} & \snd{76.5$\pm$1.1} & \best{85.1$\pm$0.4} \\
\midrule
\grp{6}{Trained on the distinctness potential}\\
\textsc{GameGate}, six-parameter policy & \trd{88.4$\pm$0.8} & \trd{53.9$\pm$0.5} & \snd{95.4$\pm$0.7} & \trd{75.6$\pm$1.5} & \snd{84.6$\pm$0.7} \\
\bottomrule
\end{tabular}

\end{table}

\FloatBarrier
\subsection{Learned versus frozen random policy}
\label{app:random_policy}
Table~\ref{tab:app_matched} separates learning the policy from merely varying the actions across nodes. At the selected \textsc{TaskGate} configuration, the frozen random policy keeps the form of Eq.~\eqref{eq:policy} but draws the five descriptor weights of each layer from $\mathcal{N}(0,0.5^2)$, with zero bias, and never trains them; each split draws its own policy, so the mean over the ten splits averages over ten random policies. The random policy improves on fixed mixing on Roman-empire, Questions, and Tolokers, the datasets where the gate helps, and lowers the score on Amazon-ratings and Minesweeper. The learned policy improves on the random one on every dataset, significantly only on Amazon-ratings and Minesweeper (Table~\ref{tab:app_matched_signif}).

Table~\ref{tab:app_seeds} repeats the comparisons on the datasets where the gate helps with five seeds per split. It tests them with the corrected resampled $t$-test of \citet{nadeau2003inference}, which allows for the overlap between the splits, and applies Holm's correction over the three datasets of each comparison. With this test, the learned policy is significantly ahead of fixed mixing on all of them. The random policy also improves on fixed mixing, significantly only on Roman-empire, and the learned policy improves on the random one, significantly on Roman-empire and with raw $p<0.05$ on Tolokers. Against a random policy with a narrower spread, closer to that of \textsc{TaskGate}, the learned policy is again significantly better on Roman-empire. Variation of the actions across nodes thus accounts for part of the gain where the gate helps, and learning adds to it.

The degree rows of Table~\ref{tab:app_matched} ask how much of the gain a degree rule recovers. A gate that sees only the degree, with two parameters per layer, is within 0.6 points of \textsc{TaskGate} on every dataset and ahead on four, significantly on Minesweeper; with five seeds, no difference between the two is significant, and the 95\% intervals lie between $-1.2$ and $+0.6$ points (Table~\ref{tab:app_seeds}). The parameter-free rule $\alpha_i=1/(1+\sqrt{d_i})$ is within 0.6 points of \textsc{TaskGate} on every dataset, with no significant difference, and $\alpha_i=1/(1+d_i)$, the self share of a mean over the node and its neighbors, within 0.4 points on all but Questions and Tolokers. Fixed mixing with $\log(1+d_i)$ as an extra input feature stays below \textsc{TaskGate} on four datasets, significantly on Roman-empire and Amazon-ratings, so the benefit comes from mixing that depends on the degree rather than from the degree as information. The symmetric normalization of GCN, $1/\sqrt{d_id_j}$ in place of the mean over the strict neighbors, also gives the neighborhood channel more weight at high-degree nodes: with it, fixed mixing is within 0.7 points of \textsc{TaskGate} on Roman-empire, Questions, and Tolokers, with no significant difference over five seeds, and significantly below it, by 1.8 points, on Amazon-ratings. A sum over the strict neighbors gives the same score as the symmetric normalization on Roman-empire but lowers fixed mixing by 4 to 12 points on the other datasets. On the heterophily suite, a degree-dependent self weight is therefore enough to recover the gain of the policy, although the actions of the full policy do not follow the degree alone (Appendix~\ref{app:where}).

Trained on the distinctness potential at the same configuration, with the gate learning rate selected by validation between $10^{-4}$ and $10^{-3}$, the policy does not improve on the random one where the random policy helps: ascent is within 0.1 points of it on Roman-empire, Questions, and Tolokers, and significantly ahead only on Amazon-ratings and Minesweeper, where the random policy loses accuracy while ascent stays at fixed mixing. Ascending and descending the potential differ significantly only on Roman-empire, where ascent is better, and on Minesweeper, where descent is better.

\begin{table}[!htbp]
\centering
\caption{Every variant at the selected \textsc{TaskGate} configuration, grouped by how the gate is trained: mean $\pm$ standard deviation over the ten splits. The frozen random policy draws the descriptor weights of each layer from $\mathcal{N}(0,0.5^2)$, with zero bias, once per split. The degree rules set $\alpha_i$ without parameters; the next row gives fixed mixing $\log(1+d_i)$ as an extra input feature, and the two rows after it replace the mean over the strict neighbors with $1/\sqrt{d_id_j}$ (symmetric normalization) or with a sum; the degree-only gate is \textsc{TaskGate} with the degree as its only descriptor. Ascent (\textsc{GameGate}) and descent use the gate learning rate selected by validation over $\{10^{-4},10^{-3}\}$.}
\label{tab:app_matched}
\begin{tabular}{lccccc}
\toprule
Model & Roman-emp. & Amazon-rat. & Minesw. & Questions & Tolokers \\
\midrule
\grp{6}{Gate not trained}\\
\textsc{Fixed} ($\alpha=1/2$) & 87.1$\pm$0.7 & \trd{53.8$\pm$0.5} & 95.3$\pm$0.5 & 74.6$\pm$1.0 & 84.2$\pm$0.6 \\
frozen random policy & 88.4$\pm$1.0 & 52.0$\pm$0.8 & 94.2$\pm$0.5 & 75.8$\pm$2.2 & 84.6$\pm$0.6 \\
degree rule $1/(1+d_i)$ & \best{89.4$\pm$0.5} & 53.6$\pm$0.5 & \trd{95.5$\pm$0.5} & 76.0$\pm$1.1 & 84.4$\pm$1.0 \\
degree rule $1/(1+\sqrt{d_i})$ & 89.0$\pm$0.4 & \trd{53.8$\pm$0.4} & \snd{95.6$\pm$0.4} & \best{77.6$\pm$1.4} & \trd{85.1$\pm$0.6} \\
\textsc{Fixed}, $\log(1+d_i)$ as input & 88.0$\pm$0.5 & 52.5$\pm$0.4 & \trd{95.5$\pm$0.5} & 76.0$\pm$2.4 & 84.4$\pm$0.9 \\
\textsc{Fixed}, symmetric normalization & 88.9$\pm$0.4 & 52.2$\pm$0.7 & \trd{95.5$\pm$0.6} & \trd{76.3$\pm$1.3} & \best{85.3$\pm$0.7} \\
\textsc{Fixed}, sum aggregation & 88.9$\pm$0.5 & 49.4$\pm$0.5 & 91.3$\pm$0.8 & 64.9$\pm$1.2 & 72.4$\pm$1.7 \\
\midrule
\grp{6}{Trained on the task loss}\\
\textsc{TaskGate} & \trd{89.2$\pm$0.4} & \best{54.0$\pm$0.3} & 95.4$\pm$0.4 & \snd{77.0$\pm$2.1} & \trd{85.1$\pm$0.5} \\
\textsc{TaskGate}, degree only & \snd{89.3$\pm$0.4} & \trd{53.8$\pm$0.6} & \best{95.7$\pm$0.5} & \best{77.6$\pm$1.2} & \snd{85.2$\pm$0.5} \\
\midrule
\grp{6}{Trained on the distinctness potential}\\
ascent (\textsc{GameGate}) & 88.4$\pm$0.8 & \snd{53.9$\pm$0.5} & 95.2$\pm$0.5 & 75.7$\pm$1.6 & 84.6$\pm$0.8 \\
descent & 87.3$\pm$0.6 & \trd{53.8$\pm$0.6} & \trd{95.5$\pm$0.5} & 75.6$\pm$0.7 & 84.4$\pm$0.4 \\
\bottomrule
\end{tabular}

\end{table}

\begin{table}[!htbp]
\centering
\caption{Paired tests for Table~\ref{tab:app_matched}, first model minus second, with the markers of Table~\ref{tab:app_signif_families} and Holm's correction within each block; comparisons with a zero paired difference use the normal approximation, as there.}
\label{tab:app_matched_signif}
\begin{tabular}{lccccc}
\toprule
Difference & Roman-emp. & Amazon-rat. & Minesw. & Questions & Tolokers \\
\midrule
\grp{6}{Against the frozen random policy}\\
\textsc{TaskGate} $-$ random & $+0.87^{\dagger}$ & $+2.01^{**}$ & $+1.11^{**}$ & $+1.25$ & $+0.43^{\dagger}$ \\
\textsc{GameGate} $-$ random & $+0.06$ & $+1.94^{**}$ & $+0.92^{**}$ & $-0.08$ & $-0.01$ \\
ascent $-$ descent & $+1.09^{**}$ & $+0.06$ & $-0.36^{**}$ & $+0.13$ & $+0.17$ \\
\midrule
\grp{6}{Against fixed mixing}\\
random $-$ \textsc{Fixed} & $+1.25^{\dagger}$ & $-1.84^{**}$ & $-1.01^{**}$ & $+1.15$ & $+0.45$ \\
\textsc{TaskGate} $-$ \textsc{Fixed} & $+2.13^{**}$ & $+0.17$ & $+0.10$ & $+2.40^{\dagger}$ & $+0.88^{**}$ \\
\textsc{GameGate} $-$ \textsc{Fixed} & $+1.32^{\dagger}$ & $+0.10$ & $-0.09$ & $+1.07$ & $+0.44$ \\
descent $-$ \textsc{Fixed} & $+0.22$ & $+0.04$ & $+0.26^{\dagger}$ & $+0.95^{\dagger}$ & $+0.27$ \\
\midrule
\grp{6}{Degree controls}\\
\textsc{TaskGate} $-$ rule $1/(1+d_i)$ & $-0.15$ & $+0.34$ & $-0.14$ & $+1.05$ & $+0.62^{\dagger}$ \\
\textsc{TaskGate} $-$ rule $1/(1+\sqrt{d_i})$ & $+0.20$ & $+0.14$ & $-0.22^{\dagger}$ & $-0.59$ & $-0.04$ \\
\textsc{TaskGate} $-$ \textsc{Fixed} with $\log(1+d_i)$ & $+1.24^{**}$ & $+1.42^{**}$ & $-0.11$ & $+1.05$ & $+0.69^{\dagger}$ \\
\textsc{TaskGate} $-$ degree-only gate & $-0.07$ & $+0.20$ & $-0.32^{**}$ & $-0.58$ & $-0.10$ \\
\midrule
\grp{6}{Normalization of the neighborhood channel}\\
\textsc{TaskGate} $-$ \textsc{Fixed}, symmetric & $+0.36$ & $+1.79^{**}$ & $-0.10$ & $+0.69$ & $-0.25$ \\
\bottomrule
\end{tabular}

\end{table}

\begin{table}[!htbp]
\centering
\caption{Five seeds per split at the selected \textsc{TaskGate} configuration: the mean paired difference over seeds and splits, first model minus second, and in brackets its 95\% interval. Each model's score on a split is its mean over the seeds (the reported seed and four more; the narrow random policy draws its weights from $\mathcal{N}(0,0.15^2)$, for an action spread closer to that of \textsc{TaskGate}). The test is the corrected resampled $t$-test of \citet{nadeau2003inference} over the ten split means, which allows for the overlap between the splits, with Holm's correction over the three datasets of each row; markers as in Table~\ref{tab:app_signif_families}.}
\label{tab:app_seeds}
\begin{tabular}{lccc}
\toprule
Difference & Roman-empire & Questions & Tolokers \\
\midrule
\textsc{TaskGate} $-$ \textsc{Fixed} & $+2.05^{**}$ [$+1.60$, $+2.49$] & $+2.05^{**}$ [$+0.81$, $+3.29$] & $+0.99^{**}$ [$+0.13$, $+1.84$] \\
\textsc{TaskGate} $-$ random & $+0.99^{**}$ [$+0.32$, $+1.65$] & $+0.85$ [$-0.17$, $+1.86$] & $+0.51^{\dagger}$ [$+0.00$, $+1.01$] \\
random $-$ \textsc{Fixed} & $+1.06^{**}$ [$+0.31$, $+1.81$] & $+1.20$ [$-0.10$, $+2.51$] & $+0.48$ [$-0.68$, $+1.65$] \\
\textsc{TaskGate} $-$ random, narrow & $+0.85^{**}$ [$+0.32$, $+1.38$] & $+0.68$ [$-1.33$, $+2.69$] & $+0.40$ [$-0.05$, $+0.84$] \\
\textsc{TaskGate} $-$ rule $1/(1+d_i)$ & $-0.12$ [$-0.60$, $+0.36$] & $+0.25$ [$-0.68$, $+1.18$] & $+0.58$ [$-0.21$, $+1.38$] \\
\textsc{TaskGate} $-$ degree-only gate & $-0.15$ [$-0.54$, $+0.24$] & $-0.31$ [$-1.13$, $+0.51$] & $-0.03$ [$-0.23$, $+0.16$] \\
\textsc{TaskGate} $-$ \textsc{Fixed}, symmetric & $+0.28$ [$-0.07$, $+0.64$] & $+0.56$ [$-1.20$, $+2.32$] & $-0.22$ [$-0.92$, $+0.48$] \\
\bottomrule
\end{tabular}

\end{table}

\FloatBarrier
\subsection{Where the gate acts}
\label{app:where}
\label{app:descriptor}
\begin{figure}[!htbp]
\centering
\includegraphics[width=0.95\textwidth]{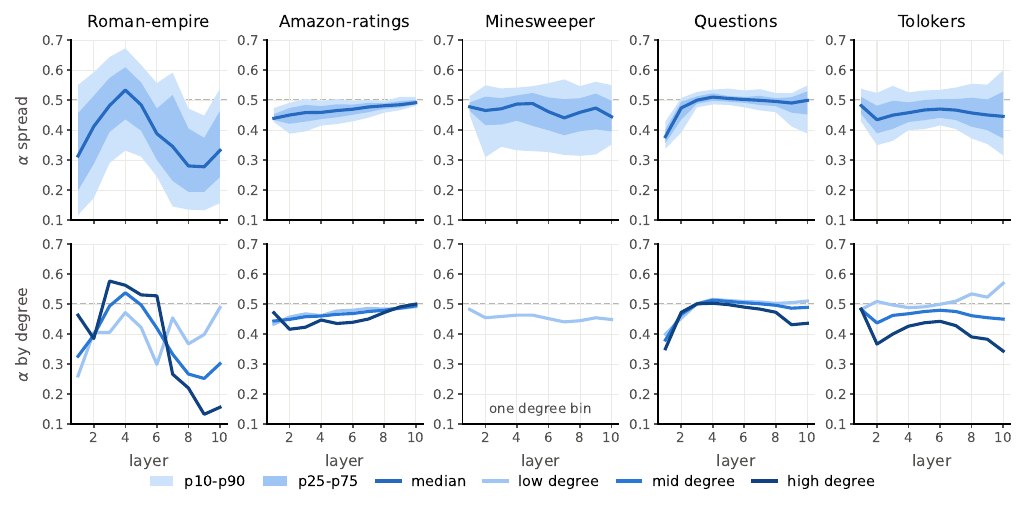}
\caption{Actions of the \textsc{TaskGate} policy layer by layer, over the nodes with at least one strict neighbor: quantiles of the per-node action $\alpha_i^\ell$ (top) and its mean within each tertile of the strict degree (bottom; Minesweeper, a nearly regular grid, has a single degree bin), averaged over the checkpoints of splits 0 to 2.}
\label{fig:app_alpha}
\end{figure}

Table~\ref{tab:app_structure} relates the actions of the \textsc{TaskGate} policy to the strict degree $d_i$ and to the local homophily $h_i$, the fraction of strict neighbors that share the node's label (used for this analysis only). Figure~\ref{fig:app_alpha} shows the actions layer by layer. On the degree-heterogeneous Questions and Tolokers, the actions follow degree, with a lower self weight at high-degree nodes, and the gate improves on fixed mixing. On Minesweeper, a regular grid, they follow local homophily even within degree bins, but the gate brings no gain. On Roman-empire, where the gain is largest, the actions vary the most and change with depth: in the early and middle layers, high-degree nodes keep more of their own channel than low-degree nodes, and in the last four layers much less, so the correlation with degree cancels in the average over layers. Local homophily plays no role there. Neither degree nor local homophily alone explains where node-wise gating helps on these five datasets.

\begin{table}[!htbp]
\centering
\caption{Where the \textsc{TaskGate} policy acts: Spearman correlation of the per-node action $\alpha_i^\ell$ with log-degree and with local homophily (also within degree bins, so that a degree rule alone cannot produce it), and the spread of the actions, averaged over layers and over the checkpoints of splits 0 to 2; the first row is the gain over \textsc{Fixed} of Table~\ref{tab:hetero}, and the last two rows are graph statistics.}
\label{tab:app_structure}
\begin{tabular}{lccccc}
\toprule
Statistic & Roman-emp. & Amazon-rat. & Minesw. & Questions & Tolokers \\
\midrule
\grp{6}{Gain of the gate}\\
\textsc{TaskGate} $-$ \textsc{Fixed} & $+1.9$ & $+0.2$ & $-0.1$ & $+1.8$ & $+1.0$ \\
\midrule
\grp{6}{Learned actions}\\
std of $\alpha$ & 0.14 & 0.03 & 0.09 & 0.04 & 0.07 \\
$\rho(\alpha, \log d_i)$ & $-0.01$ & $-0.24$ & $-0.24$ & $-0.33$ & $-0.67$ \\
$\rho(\alpha, h_i)$ & $+0.01$ & $+0.07$ & $+0.38$ & $+0.15$ & $+0.15$ \\
$\rho(\alpha, h_i)$ within degree bins & $+0.00$ & $+0.07$ & $+0.37$ & $+0.04$ & $+0.10$ \\
\midrule
\grp{6}{Graph}\\
degree coefficient of variation & 0.35 & 0.79 & 0.07 & 4.32 & 1.83 \\
edge homophily & 0.05 & 0.38 & 0.68 & 0.84 & 0.59 \\
\bottomrule
\end{tabular}

\end{table}

\FloatBarrier
\subsection{Policy transfer}
\label{app:transfer}
Table~\ref{tab:app_transfer} loads the \textsc{TaskGate} policy learned on one dataset (row, split-0 model), freezes it in the selected \textsc{TaskGate} configuration of another (column), and trains the remaining parameters on the ten splits of the target; Table~\ref{tab:free} reports the mean over the four foreign sources. On Roman-empire, Questions, and Tolokers, where the gate improves on fixed mixing, a foreign policy retains part of the gain of the target's own policy, and on Amazon-ratings and Minesweeper it stays within a point of fixed mixing. A policy trained without labels could transfer more readily, since its training signal does not depend on the source task. Table~\ref{tab:app_transfer_game} repeats the transfer with the \textsc{GameGate} policy and finds that it does not: averaged over the four sources, the label-free policy scores below the task-trained one on four targets, by 0.2 to 0.9 points, and matches it on Tolokers. The policies of Amazon-ratings and Minesweeper, trained at gate learning rate $10^{-4}$, barely move the actions from $1/2$ and stay within the variation of fixed mixing on every target, while those of the other three datasets lower accuracy on Amazon-ratings and Minesweeper.

\begin{table}[!htbp]
\centering
\caption{Policy transfer: the \textsc{TaskGate} policy learned on the source dataset (row, split-0 model) is loaded and frozen in the selected \textsc{TaskGate} configuration of the target dataset (column), and only the remaining parameters are trained, on the ten splits. The first row is the score of \textsc{Fixed} on the target; every other entry is the gain over it, in points. Gray entries freeze the target's own policy, which is not a transfer, since that policy was trained on labels of the same graph; bold marks the best transferred policy for each target. The last row is the gain of the target's own \textsc{TaskGate}, trained jointly with the backbone on every split.}
\label{tab:app_transfer}
\begin{tabular}{lccccc}
\toprule
Policy & Roman-emp. & Amazon-rat. & Minesw. & Questions & Tolokers \\
\midrule
\textsc{Fixed} trained on the target (score) & 87.3 & 53.8 & 95.4 & 75.2 & 84.1 \\
\midrule
\grp{6}{Gain over \textsc{Fixed}: frozen \textsc{TaskGate} policy from}\\
Roman-empire & \textcolor{gray}{+1.7} & $-$0.3 & $-$0.7 & +0.2 & +0.1 \\
Amazon-ratings & +1.3 & \textcolor{gray}{$-$0.2} & \best{+0.2} & +0.7 & +0.6 \\
Minesweeper & +0.8 & $-$0.1 & \textcolor{gray}{$-$0.1} & \best{+1.2} & \best{+0.7} \\
Questions & +1.0 & \best{+0.4} & 0.0 & \textcolor{gray}{+1.6} & +0.4 \\
Tolokers & \best{+1.7} & $-$0.3 & +0.1 & +1.1 & \textcolor{gray}{+1.0} \\
\midrule
\grp{6}{Gain over \textsc{Fixed}: policy trained on the target}\\
\textsc{TaskGate} & +1.9 & +0.2 & $-$0.1 & +1.8 & +1.0 \\
\bottomrule
\end{tabular}

\end{table}

\begin{table}[!htbp]
\centering
\caption{Policy transfer with the label-free policy: as Table~\ref{tab:app_transfer}, with the \textsc{GameGate} policy of the source dataset (split-0 model at the reported gate learning rate) loaded and frozen in the selected \textsc{TaskGate} configuration of the target.}
\label{tab:app_transfer_game}
\begin{tabular}{lccccc}
\toprule
Policy & Roman-emp. & Amazon-rat. & Minesw. & Questions & Tolokers \\
\midrule
\textsc{Fixed} trained on the target (score) & 87.3 & 53.8 & 95.4 & 75.2 & 84.1 \\
\midrule
\grp{6}{Gain over \textsc{Fixed}: frozen \textsc{GameGate} policy from}\\
Roman-empire & \textcolor{gray}{+1.1} & $-$1.3 & $-$0.4 & \best{+1.5} & \best{+0.7} \\
Amazon-ratings & $-$0.2 & \textcolor{gray}{$-$0.1} & \best{$-$0.2} & $-$0.1 & +0.1 \\
Minesweeper & +0.2 & \best{+0.1} & \textcolor{gray}{0.0} & $-$0.1 & +0.2 \\
Questions & +0.6 & $-$0.7 & $-$0.3 & \textcolor{gray}{+0.3} & +0.4 \\
Tolokers & \best{+0.9} & $-$0.8 & $-$0.4 & +0.5 & \textcolor{gray}{+0.6} \\
\midrule
\grp{6}{Gain over \textsc{Fixed}: policy trained on the target}\\
\textsc{TaskGate} & +1.9 & +0.2 & $-$0.1 & +1.8 & +1.0 \\
\bottomrule
\end{tabular}

\end{table}

\FloatBarrier
\subsection{Transfer to an unseen graph}
\label{app:barbell}
\begin{figure}[!htbp]
\centering
\includegraphics[width=0.95\textwidth]{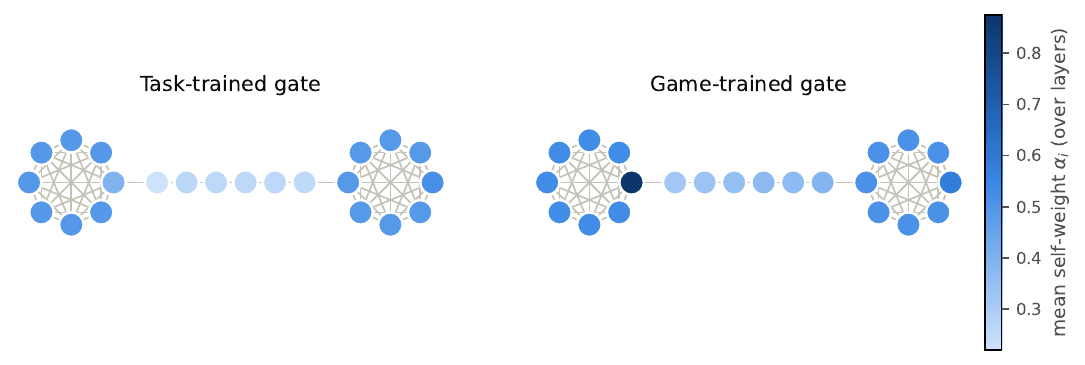}
\caption{Zero-shot transfer to an unseen barbell graph, two 8-node cliques joined by a 6-node path: mean self weight of each node over the layers, for policies trained on a synthetic shortest-path task (seed 0). Both policies give the path nodes a lower self weight than the clique nodes.}
\label{fig:app_barbell}
\end{figure}

We train \textsc{TaskGate} and \textsc{GameGate} on a synthetic shortest-path task \citep{gravina2023antisymmetric} and apply the frozen networks, without fine-tuning, to a barbell graph: two 8-node cliques joined by a 6-node path (Figure~\ref{fig:app_barbell}). In all five training seeds, both policies give the path nodes, of strict degree two, a mean self weight about $0.2$ lower than the clique nodes, of degree seven or eight: they rely more on propagation along the bridge and keep more of the self channel inside the cliques. When the degree input is removed from the descriptor, the gap disappears, so the transferred pattern comes from degree.

\FloatBarrier
\section{Training the Gate on the Distinctness Potential}
\label{app:objective}
This section complements Section~\ref{sec:objective}.

\FloatBarrier
\subsection{Gate learning rate and direction of the potential}
\label{app:dose}
\textsc{GameGate} ascends the distinctness potential with a small gate learning rate. Table~\ref{tab:app_dose} raises the gate learning rate from $10^{-4}$ to $10^{-1}$ and also reverses the training signal, descending the potential, which rewards each node for resembling its neighbors instead. Each entry is compared with fixed mixing at the same configuration.

With larger rates, the actions of ascent move toward the endpoints, as Proposition~\ref{prop:endpoint} predicts, and the mean of $\alpha(1-\alpha)$ falls from $0.25$ to below $0.1$. Except on Questions, accuracy falls as the actions harden: the largest rate is below fixed mixing, and the gain of \textsc{GameGate}, where there is one, comes from the small rate that validation selects, at which the actions stay close to $1/2$. On Questions, every rate from $10^{-3}$ on improves on fixed mixing by more than a point, but descent, which keeps the actions interior, gains about as much, so the gain there does not depend on the direction of the potential. Elsewhere, descent stays within a few tenths of a point of fixed mixing.

At the rates selected by validation, ascent is significantly better than descent only on Roman-empire and Tolokers (Table~\ref{tab:app_signif_families}). The gate weights in the last block of Table~\ref{tab:app_dose} show how the two directions differ: the ascending gate weights the self-channel norm positively and the neighborhood-channel norm negatively, so it moves each node toward whichever of its channels is larger, and the descending gate does the opposite.

\begin{table}[!htbp]
\centering
\caption{Gate learning rate and direction of the distinctness potential, at the configuration \textsc{GameGate} selects with gate learning rate $10^{-4}$. The first row is fixed mixing at that configuration; every other entry is the score minus that row, with the mean of $\alpha(1-\alpha)$ over nodes and layers at the final epoch in parentheses ($0.25$ when every action is $1/2$, $0$ when every action is $0$ or $1$). Bold marks the gate learning rate that validation selects for each direction. The last block gives the gate weights on the standardized self- and neighborhood-channel norms of the descriptor at that rate, averaged over layers and splits. On Minesweeper at $10^{-1}$, validation selects an epoch of the warm-up, before the game starts, in 5 of the 10 splits of the ascent.}
\label{tab:app_dose}
{\small
\begin{tabular}{lccccc}
\toprule
Run & Roman-emp. & Amazon-rat. & Minesw. & Questions & Tolokers \\
\midrule
\textsc{Fixed}, same configuration & 87.3$\pm$0.6 & 53.8$\pm$0.5 & 95.3$\pm$0.4 & 73.9$\pm$2.1 & 84.1$\pm$0.6 \\
\midrule
\grp{6}{Ascent (\textsc{GameGate}): score minus \textsc{Fixed} (mean $\alpha(1-\alpha)$)}\\
gate lr $10^{-4}$ & +0.3 (0.25) & \textbf{+0.1 (0.25)} & \textbf{+0.1 (0.25)} & 0.0 (0.25) & 0.0 (0.25) \\
gate lr $10^{-3}$ & \textbf{+1.1 (0.24)} & 0.0 (0.23) & $-$0.6 (0.24) & +1.8 (0.24) & \textbf{+0.5 (0.24)} \\
gate lr $10^{-2}$ & $-$0.3 (0.15) & $-$1.6 (0.11) & $-$2.5 (0.15) & \textbf{+1.8 (0.16)} & +0.3 (0.14) \\
gate lr $10^{-1}$ & $-$1.3 (0.06) & $-$1.9 (0.04) & $-$3.4 (0.08) & +1.2 (0.05) & $-$1.4 (0.05) \\
\midrule
\grp{6}{Descent: score minus \textsc{Fixed} (mean $\alpha(1-\alpha)$)}\\
gate lr $10^{-4}$ & $-$0.1 (0.25) & +0.2 (0.25) & +0.2 (0.25) & +0.6 (0.25) & $-$0.1 (0.25) \\
gate lr $10^{-3}$ & \textbf{0.0 (0.25)} & +0.1 (0.24) & \textbf{0.0 (0.25)} & \textbf{+1.7 (0.25)} & +0.1 (0.25) \\
gate lr $10^{-2}$ & +0.1 (0.23) & \textbf{+0.2 (0.17)} & $-$0.1 (0.23) & +1.6 (0.23) & \textbf{+0.1 (0.21)} \\
gate lr $10^{-1}$ & $-$0.2 (0.24) & +0.6 (0.19) & 0.0 (0.24) & +0.9 (0.24) & 0.0 (0.16) \\
\midrule
\grp{6}{Gate weights on the self / neighborhood channel norms, at the selected rate}\\
ascent (\textsc{GameGate}) & +0.15 / $-$0.18 & +0.03 / 0.00 & +0.02 / $-$0.02 & +1.01 / $-$0.98 & +0.15 / $-$0.11 \\
descent & $-$0.11 / +0.13 & $-$0.26 / +0.44 & $-$0.12 / +0.16 & $-$0.12 / +0.12 & $-$0.18 / +0.37 \\
\bottomrule
\end{tabular}
}
\end{table}

\FloatBarrier
\subsection{Scarce labels}
\label{app:scarce}
A label-free gate could matter most when labels are few. We train every variant at its selected configuration, \textsc{GameGate} at its selected gate learning rate, on 20\% or 5\% of each training split, the same subset for all variants, with the validation and test sets unchanged (Table~\ref{tab:app_scarce}). The label-free gate is ahead of the task-trained one only where the task-trained gate falls below fixed mixing. On Roman-empire and Questions, the task-trained gate gains more over fixed mixing with 20\% or 5\% of the labels than with all of them, and \textsc{GameGate} gains less but still improves on fixed mixing. On Tolokers and Minesweeper, the task-trained gate falls below fixed mixing at both label fractions. \textsc{GameGate} stays at fixed mixing on Tolokers at both fractions and on Minesweeper with 5\% of the labels, where validation selects an epoch of the warm-up, before the gate starts to play, in 5 of the 10 splits; on Minesweeper with 20\% of the labels, it falls below fixed mixing too. On Amazon-ratings, the three variants stay within half a point of each other. The best classic backbone, trained on the same subsets, gains more over fixed mixing than \textsc{TaskGate} only on Tolokers and about as much on Amazon-ratings; with 5\% of the labels, it falls well below fixed mixing on Roman-empire and Minesweeper.

\begin{table}[!htbp]
\centering
\caption{Scarce labels: each variant at its selected configuration, trained on 20\% or 5\% of each official training split (the same subset for every variant; validation and test sets unchanged). Mean $\pm$ standard deviation over the ten splits; small numbers give the difference from \textsc{Fixed} at the same label fraction. Best classic is the backbone of Table~\ref{tab:hetero} at its selected configuration.}
\label{tab:app_scarce}
\begin{tabular}{lccccc}
\toprule
Model & Roman-emp. & Amazon-rat. & Minesw. & Questions & Tolokers \\
\midrule
\grp{6}{20\% of the training labels}\\
\textsc{Fixed} & 71.9$\pm$0.7 & 41.1$\pm$0.8 & 91.5$\pm$0.6 & 68.8$\pm$1.6 & 80.7$\pm$0.7 \\
\textsc{TaskGate} & 75.2$\pm$0.9\,{\scriptsize$+3.2$} & 41.3$\pm$0.7\,{\scriptsize$+0.2$} & 90.6$\pm$0.7\,{\scriptsize$-0.9$} & 73.2$\pm$1.0\,{\scriptsize$+4.4$} & 79.5$\pm$1.8\,{\scriptsize$-1.1$} \\
\textsc{GameGate} & 73.1$\pm$1.2\,{\scriptsize$+1.2$} & 41.2$\pm$0.7\,{\scriptsize$+0.1$} & 89.5$\pm$1.1\,{\scriptsize$-1.9$} & 70.8$\pm$2.4\,{\scriptsize$+2.0$} & 80.8$\pm$1.3\,{\scriptsize$+0.1$} \\
Best classic & 73.2$\pm$0.5\,{\scriptsize$+1.3$} & 41.4$\pm$0.9\,{\scriptsize$+0.3$} & 90.2$\pm$0.9\,{\scriptsize$-1.2$} & 72.4$\pm$2.9\,{\scriptsize$+3.6$} & 81.6$\pm$1.5\,{\scriptsize$+0.9$} \\
\midrule
\grp{6}{5\% of the training labels}\\
\textsc{Fixed} & 58.9$\pm$0.9 & 36.5$\pm$0.7 & 83.9$\pm$3.5 & 65.7$\pm$2.9 & 75.8$\pm$1.0 \\
\textsc{TaskGate} & 61.1$\pm$1.3\,{\scriptsize$+2.2$} & 37.0$\pm$0.8\,{\scriptsize$+0.5$} & 81.8$\pm$2.3\,{\scriptsize$-2.1$} & 68.9$\pm$2.5\,{\scriptsize$+3.2$} & 73.4$\pm$1.8\,{\scriptsize$-2.3$} \\
\textsc{GameGate} & 59.4$\pm$1.2\,{\scriptsize$+0.5$} & 36.8$\pm$0.3\,{\scriptsize$+0.4$} & 83.6$\pm$2.2\,{\scriptsize$-0.3$} & 68.7$\pm$2.4\,{\scriptsize$+3.0$} & 75.6$\pm$1.6\,{\scriptsize$-0.2$} \\
Best classic & 52.2$\pm$1.8\,{\scriptsize$-6.7$} & 36.4$\pm$0.7\,{\scriptsize$-0.1$} & 72.1$\pm$6.0\,{\scriptsize$-11.7$} & 68.5$\pm$3.0\,{\scriptsize$+2.9$} & 77.4$\pm$2.5\,{\scriptsize$+1.6$} \\
\bottomrule
\end{tabular}

\end{table}
\FloatBarrier
\subsection{Unilateral regret of the trained gates}
\label{app:probe}
With the channels fixed, the distinctness game of Section~\ref{sec:game} has pure Nash equilibria, in which every nondegenerate node takes an endpoint action (Proposition~\ref{prop:endpoint}). \textsc{GameGate} does not choose the actions node by node: it trains one policy that sets all of them, so training does not guarantee that the actions reach an equilibrium (Proposition~\ref{prop:amortized}). This appendix measures how close the trained actions get.

\paragraph{Measure.}
We use the regret of each player: the payoff it could gain by changing only its own action, while every other action stays fixed. A profile is a Nash equilibrium exactly when every regret is zero, and the sum of the regrets is the NashConv measure used to track equilibrium learning in multiagent systems \citep{lanctot2017unifiedgametheoreticapproachmultiagent}. We freeze a trained network and, at layer $\ell$, write the payoff of node $i$ as a function of its own action,
\[
 f_i(a)=u_i^\ell(a,\boldsymbol{\alpha}_{-i}^\ell)=A_i a^2+B_i a+C_i,
 \qquad A_i\geq0.
\]
By Proposition~\ref{prop:endpoint} the best response is $0$ or $1$, so the regret is exact:
\[
 \operatorname{Regret}_i=\max\{f_i(0),f_i(1)\}-f_i(\alpha_i^\ell).
\]
Raw regrets scale with the channel values, so we divide each regret by the node's payoff range on $[0,1]$, which places it in $[0,1]$, and average over the nodes and layers of the network. Unlike NashConv, this normalized average does not grow with the number of players, so it can be compared across graphs.

\paragraph{Control.}
Regret can fall as the actions approach $0$ or $1$, even when nodes receive the wrong endpoint. With $D_i=f_i(1)-f_i(0)$,
\[
 \operatorname{Regret}_i
 =A_i\alpha_i^\ell(1-\alpha_i^\ell)
 +(1-\alpha_i^\ell)[D_i]_+
 +\alpha_i^\ell[-D_i]_+,
 \qquad [z]_+=\max\{z,0\},
\]
where the first term vanishes at either endpoint and the other two only at the better one. We therefore compare the trained actions with the same actions permuted at random among the nodes of each layer. The permutation keeps how close the actions are to the endpoints and changes only which node receives which action.

\paragraph{Protocol.}
We evaluate the checkpoints of splits 0 to 2 at the epoch that validation selects: \textsc{TaskGate}, the frozen random policy, and \textsc{Fixed} at their selected configurations, and \textsc{GameGate} at each gate learning rate. The players of a layer are the nodes with at least one strict neighbor; nodes whose two channels coincide ($A_i=0$, up to a relative tolerance of $10^{-12}$) have a constant payoff and are left out.

\paragraph{Results.}
Table~\ref{tab:app_regret} reports the regret of the trained actions and, in parentheses, of their permutation. With every action at $1/2$, as in \textsc{Fixed}, the regret is about $0.9$. Under \textsc{GameGate}, regret decreases as the gate learning rate grows. At $10^{-4}$ the actions stay near $1/2$, and the trained and permuted actions have the same regret. At $10^{-1}$ the actions are almost binary, and the trained actions have lower regret than their permutation on every dataset where they are measured (0.30 to 0.36, against 0.43 to 0.67): the policy gives the nodes their better endpoint more often than a random assignment would. At the rates that validation selects, the difference is small, except on Questions. The actions of \textsc{TaskGate} are about as far from an equilibrium as their permutation, or farther, and slightly farther than uniform actions: the task-trained gate does not follow the incentives of the distinctness game.

\begin{table}[!htbp]
\centering
\caption{Relative unilateral regret in the distinctness game (lower is closer to a pure Nash equilibrium): the regret of each player divided by its payoff range on $[0,1]$, averaged over players, layers, and the checkpoints of splits 0 to 2, for the deployed actions and, in parentheses, for the same actions permuted among the players of each layer. The second block gives the mean $\alpha(1-\alpha)$ of the deployed actions ($0.25$ when every action is $1/2$, $0$ when every action is $0$ or $1$). On Minesweeper at gate learning rate $10^{-1}$, validation selects an epoch before the first game update in all three splits, so no game actions are measured (--).}
\label{tab:app_regret}
\begin{tabular}{lccccc}
\toprule
Policy & Roman-emp. & Amazon-rat. & Minesw. & Questions & Tolokers \\
\midrule
\grp{6}{Relative regret of the deployed actions (the same actions permuted among the players)}\\
\textsc{Fixed} ($\alpha=1/2$) & 0.92 (0.92) & 0.93 (0.93) & 0.93 (0.93) & 0.87 (0.87) & 0.93 (0.93) \\
frozen random policy & 0.89 (0.83) & 0.87 (0.84) & 0.89 (0.85) & 0.81 (0.85) & 0.88 (0.84) \\
\textsc{TaskGate} & 0.94 (0.89) & 0.95 (0.94) & 0.94 (0.92) & 0.94 (0.95) & 0.94 (0.94) \\
\textsc{GameGate}, gate lr $10^{-4}$ & 0.94 (0.95) & 0.93 (0.93) & 0.95 (0.95) & 0.93 (0.94) & 0.94 (0.94) \\
\textsc{GameGate}, gate lr $10^{-3}$ & 0.91 (0.94) & 0.88 (0.89) & 0.93 (0.94) & 0.89 (0.91) & 0.92 (0.94) \\
\textsc{GameGate}, gate lr $10^{-2}$ & 0.61 (0.69) & 0.67 (0.66) & 0.87 (0.89) & 0.72 (0.84) & 0.61 (0.67) \\
\textsc{GameGate}, gate lr $10^{-1}$ & 0.34 (0.48) & 0.36 (0.43) & -- & 0.33 (0.67) & 0.30 (0.53) \\
\midrule
\grp{6}{Mean $\alpha(1-\alpha)$}\\
\textsc{Fixed} ($\alpha=1/2$) & 0.25 & 0.25 & 0.25 & 0.25 & 0.25 \\
frozen random policy & 0.20 & 0.21 & 0.21 & 0.22 & 0.21 \\
\textsc{TaskGate} & 0.21 & 0.25 & 0.24 & 0.24 & 0.24 \\
\textsc{GameGate}, gate lr $10^{-4}$ & 0.25 & 0.25 & 0.25 & 0.25 & 0.25 \\
\textsc{GameGate}, gate lr $10^{-3}$ & 0.24 & 0.23 & 0.24 & 0.25 & 0.24 \\
\textsc{GameGate}, gate lr $10^{-2}$ & 0.15 & 0.15 & 0.23 & 0.20 & 0.14 \\
\textsc{GameGate}, gate lr $10^{-1}$ & 0.06 & 0.05 & -- & 0.05 & 0.05 \\
\bottomrule
\end{tabular}

\end{table}


\FloatBarrier
\section{Computational Cost}
\label{app:cost}

Table~\ref{tab:app_cost} gives the cost of one layer of each \method{} variant at width $d=256$ on the Roman-empire graph, on one NVIDIA A100 40GB: the trainable parameters that one more layer adds, and the median time of a forward pass and of a training step (forward and backward), each measured as the increase per added layer. A layer has the self and neighborhood transforms, with $2(d^2+d)$ parameters, the linear map of its residual connection, with $d^2+d$, and batch normalization, with $2d$, so a layer with fixed mixing has $3d^2+5d=197{,}888$ parameters. The gate adds one parameter for \textsc{Layer}, $5+1=6$ for \textsc{TaskGate} and \textsc{GameGate}, and $(5+1)\cdot8+(8+1)=57$ for \textsc{Task-MLP}. The node-wise policy lengthens the training step by about a sixth over fixed mixing, and the game update of \textsc{GameGate} roughly doubles it.

\begin{table}[!htbp]
\centering
\caption{Cost of one \method{} layer at width 256 on Roman-empire: trainable parameters, and median forward and training-step times in milliseconds, each measured as the increase per added layer. For \textsc{GameGate}, the game update is given in parentheses.}
\label{tab:app_cost}
\begin{tabular}{lrrr}
\toprule
Variant & Parameters & Forward (ms) & Training step (ms) \\
\midrule
\textsc{Fixed} & 197,888 & 1.36 & 3.95 \\
\textsc{Layer} & 197,889 & 1.34 & 4.11 \\
\textsc{TaskGate} & 197,894 & 1.78 & 4.57 \\
\textsc{Task-MLP} & 197,945 & 1.82 & 4.66 \\
\textsc{GameGate} & 197,894 & 1.78 & 4.57 (+4.07 game) \\
\bottomrule
\end{tabular}

\end{table}

\end{document}